\documentclass[journal]{IEEEtran}

\usepackage{cite}

\ifCLASSINFOpdf
\usepackage[pdftex]{graphicx}
\else
\usepackage[dvips]{graphicx}
\fi

\usepackage{amsmath}
\usepackage{amssymb}
\usepackage{multirow}
\usepackage{caption}
\usepackage{subcaption}

\begin{document}

\title{Knowledge-Routed Visual Question Reasoning: Challenges for Deep Representation Embedding}

\author{Qingxing Cao, Bailin Li, Xiaodan Liang, Keze Wang, and Liang Lin
	\IEEEcompsocitemizethanks{
		\IEEEcompsocthanksitem Q.Cao and X. Liang are with the School of Intelligent Systems Engineering, Sun Yat-sen University, China. B. Li is with DarkMatter AI Research. K. Wang is with Department of statistics, University of California, Los Angeles, California 90095. L. Lin is with the School of Computer Science and Engineering, Sun Yat-sen University, Key Laboratory of Machine Intelligence and Advanced Computing, Ministry of Education Engineering Research Center for Advanced Computing Engineering Software of Ministry of Education, China.
		}}
% The paper headers
\markboth{IEEE Transactions on Neural Networks and Learning Systems, ~Vol.~X, No.~X, XXX}%
{Shell \MakeLowercase{\textit{et al.}}: Bare Demo of IEEEtran.cls for IEEE Journals}
\maketitle

% As a general rule, do not put math, special symbols or citations
% in the abstract or keywords.
\begin{abstract}
Though beneficial for encouraging the Visual Question Answering (VQA) models to discover the underlying knowledge by exploiting the input-output correlation beyond image and text contexts, the existing knowledge VQA datasets are mostly annotated in a crowdsource way, e.g., collecting questions and external reasons from different users via the internet. In addition to the challenge of knowledge reasoning, how to deal with the annotator bias also remains unsolved, which often leads to superficial over-fitted correlations between questions and answers.
To address this issue, we propose a novel dataset named Knowledge-Routed Visual Question Reasoning for VQA model evaluation. 
Considering that a desirable VQA model should correctly perceive the image context, understand the question, and incorporate its learned knowledge, our proposed dataset aims to cutoff the shortcut learning exploited by the current deep embedding models and push the research boundary of the knowledge-based visual question reasoning.  
Specifically, we generate the question-answer pair based on both the Visual Genome scene graph and an external knowledge base with controlled programs to disentangle the knowledge from other biases. The programs can select one or two triplets from the scene graph or knowledge base to push multi-step reasoning, avoid answer ambiguity, and balanced the answer distribution. 
In contrast to the existing VQA datasets, we further imply the following two major constraints on the programs to incorporate knowledge reasoning: i) multiple knowledge triplets can be related to the question, but only one knowledge relates to the image object. 
This can enforce the VQA model to correctly perceive the image instead of guessing the knowledge based on the given question solely; ii) all questions are based on different knowledge, but the candidate answers are the same for both the training and test sets. We make the testing knowledge unused during training to evaluate whether a model can understand question words and handle unseen combinations. 
Extensive experiments with various baselines and state-of-the-art VQA models are conducted to demonstrate that there still exists a big gap between the model with and without groundtruth supporting triplets when given the embedded knowledge base. This reveals the weakness of the current deep embedding models on the knowledge reasoning problem.
\end{abstract}

% Note that keywords are not normally used for peerreview papers.

% For peer review papers, you can put extra information on the cover
% page as needed:
% \ifCLASSOPTIONpeerreview
% \begin{center} \bfseries EDICS Category: 3-BBND \end{center}
% \fi
%
% For peerreview papers, this IEEEtran command inserts a page break and
% creates the second title. It will be ignored for other modes.
\IEEEpeerreviewmaketitle

\begin{figure}[t]
	\centering
	\includegraphics[width=0.95\linewidth]{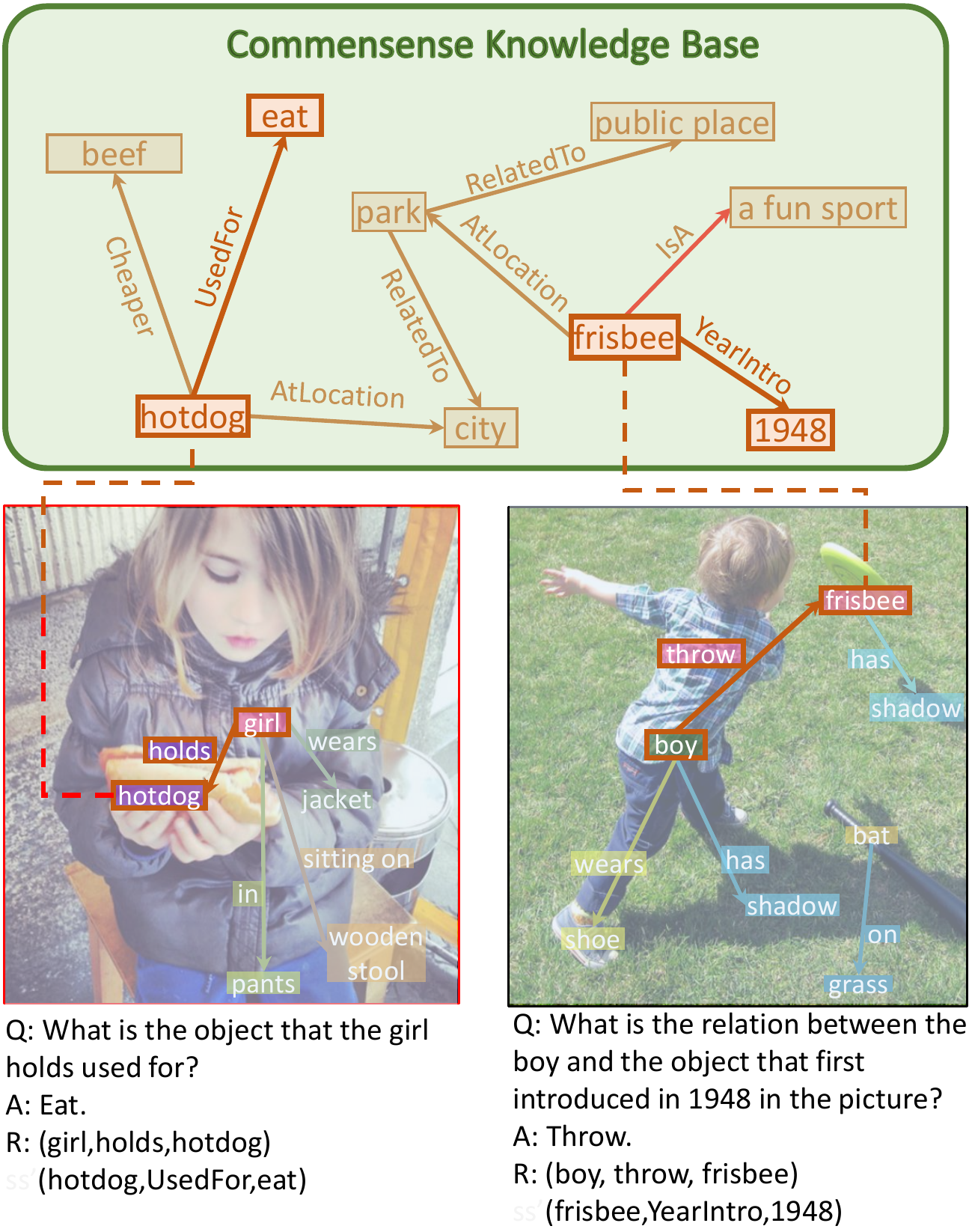}
	\caption{Examples of our proposed KRVQR dataset. Each question is based on one or two triplets selected from the scene graph annotations or external commonsense knowledge base. To answer the question, a desirable VQA model must perform multi-step reasoning over both the image context and background knowledge.}
	\label{fig:sample}
\end{figure}

\section{Introduction}

As a core task towards the complete AI, Visual Question Answering (VQA) demands the understanding of not only image and text but also the background knowledge that beyond observable inputs~\cite{VQA}. For example, to correctly answer the question ``what is the object that the girl is holding used for'', an agent should understand the reasoning process, perceive the ``girl'' and the object with relation ``holding'', and use the background knowledge ``hotdog UsedFor eat'' to generate the correct answer.

To push the VQA models to correctly understand the image and text, a balanced VQA benchmark plays a crucial role. However, the earlier VQA datasets often contain the annotator bias, which is exploited by VQA models to guess the answer instead of understanding the image and questions. To remove such superficial over-fitted relations, a lot of efforts have been made to balance the distribution of image/question and answer~\cite{balanced_vqa_v2, clevr, GQA}, and have lead to great success on the fused image and text representation for answer prediction~\cite{MLB, film, mcan, lxmert}.

To enforce the VQA models to discover the knowledge by relations between questions and answers, current knowledge VQA datasets still crowdsourcingly annotate a large amount of the question-answer pairs to implicitly incorporate the knowledge. Actually, without applying no proper control, the statistics and annotator biases may be displayed throughout the dataset and result in the superficial relations between the questions and answers. Besides, the knowledge relation is inevitably tangled with other superficial relations due to the vast range of knowledge and relatively small annotated samples.
These datasets contain the shortcut between the question and the answer. Instead of encouraging the existing VQA models to learn to perform knowledge reasoning, these datasets push them to increase the model's capacity and consume a large-scale of extra data to resource-consumingly fit all kinds of superficial correlation. 

% On the contrary, humans can fit into a new environment by retrieving the related knowledge efficiently without such a resource consumable process.
%A desirable VQA model should be able to correctly perceive visual evidence in the image, understand the reasoning logic in the question, and incorporate the supporting knowledge. 

Attempt to overcome the above-mentioned limitation of the existing datasets, we propose a novel challenging Knowledge-Routed Visual Question Reasoning (KRVQA) dataset to correctly evaluate the following abilities of a VQA model: i) correctly perceiving visual evidence in the image, ii) understanding the reasoning logic in the question, and iii) incorporating the supporting knowledge, by further disentangling the knowledge from other superficial relations.
Similar to the previous unbiased VQA datasets~\cite{clevr,GQA}, we generate the question-answer pairs based on tight controlled programs operates on the Visual Genome scene graph~\cite{VG} and an external knowledge base~\cite{FVQA}. Each program consists of one or two triplets from the scene graph or knowledge base to ensure the questions can be and must be answered with the provided knowledge. 
%Since the knowledge can be represented by various text, we provide the knowledge base to VQA models to avoid answering ambiguity and to ensure the model can obtain enough background knowledge for answering the questions.
Besides the same constraints on scene graph triplets, we further imply the following constraints on the knowledge triplets. Firstly, if an answer is from the knowledge base, then only one knowledge triplet is related to the image, such that this sample has a unique answer. Secondly, if an answer is from the image, then there should be multiple knowledge triplets related to the question, but only one knowledge relates to the objects in the image. We imply this constraint to prevent a model from memorizing the knowledge to predict an answer without perceiving the image context. Thirdly, all programs must rely on different knowledge triplets. Thus, a VQA model must learn to reason on the knowledge base. We enforce these constraints such that the model must correctly parse the question instead of memorizing the mapping from question to knowledge. Lastly, we balance the knowledge answer distribution to prevent the model from guessing the answer, but relax the image answer distribution to enforce the model to recognize small and rarely appeared objects. Given an image, we merge its scene graph with the knowledge base to form an image-specific knowledge graph, then we search the triplets that meet the above constraints and compose the program to generate the question-answer pairs. We illustrate the sample of the generated question-answer in Figure~\ref{fig:sample}.

We conduct extensive experiments on various knowledge embedding baselines and state-of-the-art VQA models.
We have shown that even with the state-of-the-art knowledge graph embedding method which obtains over $99\%$ accuracy on knowledge triplets prediction, the current state-of-the-art VQA models still achieve low answering accuracy on our proposed KRVQR dataset. These experimental results demonstrate that our proposed dataset poses a new challenge towards current black-box VQA models and can push the boundary of visual knowledge reasoning.

\begin{table}[t!]
	\centering
	\caption{Comparison of KRVQR and previous datasets. Our KRVQR dataset is the first large-scale (more than 100,000 QA pairs) dataset that requires knowledge reasoning on natural images.}
	\resizebox{\columnwidth}{!}{
		\begin{tabular}{lcccc}
			\hline
			Dataset&Large&Knowledge&Supporting &Diagnostic\\
			&Scale&Reasoning&Knowledge Base&\\
			\hline
			CLEVR~\cite{clevr} & \checkmark & & & \checkmark\\
			VQAv2~\cite{balanced_vqa_v2} & \checkmark & & & \\
			GQA~\cite{GQA}  & \checkmark &  & & \checkmark\\
			FVQA~\cite{FVQA}  & & \checkmark & \checkmark & \checkmark \\
			OKVQA~\cite{okvqa}  & & \checkmark & & \\
			VCR~\cite{vcr}  & \checkmark & \checkmark & &\\
			\hline
			Our KRVQR  & \checkmark & \checkmark & \checkmark & \checkmark\\
			\hline
		\end{tabular}
	}
	\label{tab:datasets}
\end{table}

\begin{table*}[!t]
	\centering
	\caption{Templates for the question-answer pair generation.}
% 	\resizebox{0.85\textwidth}{!}{
		\begin{tabular}{|c|c|c|c|c|}
			\hline
			Step & Qtype & Question Semantics & Answer & Reason\\
			\hline
			\multirow{3}{*}{1} & 0 & What is the relationship of $<$A$>$ and $<$B$>$? & $<$R$>$ & \multirow{3}{*}{(A, R, B)}\\
			& 1 & what is $<$A$>$ $<$R$>$? & $<$B$>$ &\\
			& 2 & what $<$R$>$ $<$B$>$? & $<$A$>$ &\\
			\hline
			\multirow{4}{*}{1} & 3 & What is the relation $<$of$>$ the object that $<$A$>$ $<$R1$>$ and $<$C$>$? & $<$R2$>$ & \multirow{4}{*}{}\\
			& 4 & What is the relation of $<$A$>$ and the object that $<$R2$>$ $<$C$>$? & $<$R1$>$ &(A, R1, B) \\
			& 5 & what $<$A$>$ $<$R1$>$ $<$R2$>$? & $<$C$>$ &(B, R2, C)\\
			& 6 & what $<$R1$>$ $<$R2$>$ $<$C$>$? & $<$A$>$ &\\
			\hline
		\end{tabular}
% 	}
	\label{tab:templates}
\end{table*}

\section{Related Works}
\subsection{Visual question answering} 
The Visual Question Answering(VQA) task is to predict the answer given the image and question. It requires co-reasoning over multiple domains, including image and text and background knowledge.
The pioneer works used the CNN-LSTM-based architecture and trained the neural networks in an end-to-end manner. Later, the attention mechanism~\cite{DBLP:journals/corr/IlievskiYF16, Xu2016,shih2016att, Visual7W, san, HiCoAtt} has been shown as a strong method in terms of the answering accuracy on various VQA benchmarks.
Subsequently, a lot of efforts have been devoted to the joint embedding of image and question representation~\cite{MCB, MLB, MFB, MUTAN}, The combination of the attention mechanism and compact bilinear multimodal fusion has further improved performance~\cite{Anderson2017up-down,vqav2winner}. Most recently, BERT-like architecture~\cite{mcan,vl-bert,lxmert} has been widely used and pretrained on various image-text tasks like image captioning and visual question answering to embed the text token and image object simultaneously. Moreover, some works~\cite{comet,kbert} have trained the BERT-like model on large scale corpus to encode the commonsense knowledge.

Besides the end-to-end neural networks, other works explore different ways of addressing compositional visual reasoning and interpretable answer prediction. The neural modular networks~\cite{nmn,lnmn,e2emn} have attracted substantial interest. Instead of inferencing on fixed architecture, these works exploited sub-modules to solve subtasks and assemble their results for final answer prediction.
Most recently, graph convolution networks and symbolic inference have been widely studied. For instance, 
\cite{norcliffebrown2018learning} applied a graph convolution network to obtain the question-specific graph representation and interactions in images. \cite{NeuralSymbolic} transformed images to scene graphs and questions to functional layouts and then performed symbolic inference on the graph. 

\subsection{VQA based on external knowledge} 
As the VQA requires understanding the knowledge, other strands of research have attempted to leverage such information beyond image-question pairs. These works either retrieve the external common knowledge and basic factual knowledge to answer the questions~\cite{externVQA, FVQA, Narasimhan_2018_ECCV} or actively obtaining extra information and predicting the answer~\cite{zhu2017cvpr, Misra_2018_CVPR, IQA}.
The proposed model in FVQA~\cite{FVQA} answered the questions by learning query mappings and retrieve information in the knowledge base. \cite{Narasimhan_2018_ECCV} answers the questions by predicting the key triplet related to the question. These two methods require groundtruth triplets as the supervision signal and are only evaluated on one-hop reasoning questions.

\cite{Li_2018_CVPR} tried to learn the complementary relationship between questions and answers and introduced question generation as a dual-task to improve VQA performance. ~\cite{Teney_2018_ECCV} dynamically selected example questions and answers during the training and encoded examples into a support set for answering the questions. \cite{narasimhan2018out} also utilized a graph convolution network but embedded the retrieved knowledge with image representation as a node, and they evaluated their method on the FVQA~\cite{FVQA} knowledge-based VQA dataset.

\subsection{VQA datasets}
There are lots of VQA datasets that have been proposed in recent years and the VQA~\cite{VQA,balanced_vqa_v2} are the most popular VQA datasets. However, the annotation process does not involve hard controls and the dataset contains a severe bias, which is exploited by various black-box models to attain state-of-the-art performance~\cite{balanced_vqa_v2, ECCV16baseline}. This concern has led to the proposal of unbiased datasets~~\cite{clevr,CountQA,IQA,DVQA,GQA} that generate the question with programs to control the distribution of both questions and answers.

Recently, lots of works have focused on incorporating external knowledge for visual question answering.
FVQA~\cite{FVQA} constructs a knowledge base by collecting the knowledge triplets from WebChild~\cite{webchild}, ConceptNet~\cite{conceptnet}, and DBpedia~\cite{dbpedia}. Then it extracts the visual concepts and matched image with the knowledge to annotate the question-answer pairs. OKVQA~\cite{okvqa} manually annotates the questions that require some outside knowledge. VCR~\cite{vcr} crowdsourcingly annotates not only the question-answer pair, but also annotates the rationale of the answer. The VCR casts both the answer and rationale prediction as a four-way multiple-choice problem. This work further applies adversarial matching to select the most confusing rationale from other samples as choice candidates.
However, the manual annotation process inevitably includes the annotator bias. Also, annotating the rationale as a sentence introduces the extra difficulty in language understanding, making it hard to control the supporting knowledge range.
To address the aforementioned issue, our proposed KRVQR leverages a knowledge base to limit the boundary of candidate knowledge, and ensures the question-answer relations is based on a known knowledge triplet. Compared with previous knowledge VQA datasets, the proposed KRVQR advances in not only minimizing the dataset bias, but also providing groundtruth knowledge for further inspection.

\section{Knowledge-Routed Visual Question Reasoning dataset}
The proposed Knowledge-Routed Visual Question Reasoning(KRVQA) dataset aims to evaluate the reasoning ability of external knowledge while prevent the models from exploiting superficial relations. It consists of $32,910$ images and $157,201$ question-answer pairs of different types for testing a variety of skills of the given VQA model.
Compared with previous datasets, our proposed KRVQA has the following advantages: i) We use the tightly controlled program on explicit scene graph and knowledge base to minimize the bias on both image and knowledge; ii) We provide the groundtruth program to enable inspection on the reasoning behavior of a VQA model; and iii) each knowledge triplet only appears once for all questions to pose the new challenges for deep embedding networks to correctly embed the knowledge base and handle the unseen questions. Detailed comparisons on the properties of our KRVQA dataset with existing datasets can be found in Table~\ref{tab:datasets}.

We construct the KRVQA dataset based on scene graph annotations from Visual Genome~\cite{VG} dataset and the knowledge base from FVQA~\cite{FVQA}. To generate the unbiased question-answer pairs, we first clean the object and relation name in the Visual Genome scene graph annotations. Then given an image, we combine its scene graph with the related knowledge triplets to form the image-specific knowledge graph. We further extract the facts from the graph and compose them into a reasoning program. Lastly, we generate the question-answer pairs based on the program layout and predefined question templates.

\subsection{Image-specific Knowledge Graph}
Given an image, we first clean its scene graph and integrate the scene graph with an external knowledge base to form an image-specific knowledge graph, which depicts the objects, relations, and knowledge related to this image.

Specifically, given the image and its corresponding scene graph, we only keep the objects and relations that have the ``synset'' annotations and use the ``synset'' as their class label to reduce the semantic ambiguity. For instance, all objects with the name ``bike'' or ``bicycle'' have the same synset ``bicycle'' as the label.
To ensure the generated questions require reasoning ability on external knowledge, we incorporate the large-scale knowledge base from FVQA~\cite{FVQA}. The knowledge base has a total of $193,449$ facts from WebChild~\cite{webchild}, ConceptNet~\cite{conceptnet}, and DBpedia~\cite{dbpedia}. Each fact consists of three entries: head entity, relation, and tail entity.
This knowledge base is merged the image scene graph by matching the knowledge entries with object synsets. The scene graph extended by the matched knowledge triplets is the image-specific knowledge graph, which is used to generate the question-answer pairs in our proposed KRVQA dataset.

\subsection{Triplets sampling and entry query}
The question with single-step inference, such as ``What is on the desk?'', can't evaluate the compositional reasoning ability and drive the VQA model to reason on image and knowledge base simultaneously. Thus, our KRVQA generates diverse multi-step reasoning questions via the compositional program based on a series of correlated triplets, such as the question ``what is the object that the girl is holding used for''. 

Specifically, given the image-specific knowledge graph, we perform traverse on the graph and randomly sample the traversed route with a maximum length of $2$ to generate continuously linked triplets like ``$(a, r_1, b)-(b, r_2, c)$''.

We then sample one of the six elementary queries on each sample triplet to generate questions that require different types of inference. Given a triplet``(a, r, b)'', where the ``a'', ``r'' and ``b'' represent \textit{head entity}, \textit{relation}, and \textit{tail entity} respectively, each query type has the following semantics:
\begin{itemize}
	\item[-] \textbf{$Q_{ab\_I}$}: Given a \textit{head entity} and an \textit{tail entity}, return the \textit{relation} between the \textit{head entity} and the \textit{tail entity} exploited in the image.
	\item[-] \textbf{$Q_{ar\_I}$}: Given a \textit{head entity} and an \textit{relation}, return the \textit{tail entity} exploited in the image.
	\item[-] \textbf{$Q_{rb\_I}$}: Given an \textit{relation} and an \textit{tail entity}, return the \textit{head entity} exploited in the image.
	\item[-] \textbf{$Q_{ab\_K}$}: Given a \textit{head entity} and an \textit{tail entity}, return the \textit{relation} between the \textit{head entity} and the \textit{tail entity} in the knowledge base.
	\item[-] \textbf{$Q_{ar\_K}$}: Given a \textit{head entity} and an \textit{relation}, return the \textit{tail entity} in the knowledge base.
	\item[-] \textbf{$Q_{rb\_K}$}: Given an \textit{relation} and an \textit{tail entity}, return the \textit{head entity} in the knowledge base. 
\end{itemize}

\subsection{Program Layout}
With the sampled queries and triplets, we compose them into a program layout. For example, the sampled ``(girl, holds, hotdog)'' with query $Q_{ar\_I}$ is represented as ($Q_{ar\_I}$, girl, holds), which requires inferring the ``hotdog'' given the ``girl'' and ``holds'' in the image scene graph. And the second sampled query is ``(${Q_{ar\_K}}$, hotdog, UsedFor)'' that requires inferring the ``eat'' given the head entity and relation in the knowledge base.
we compose them into ``(${Q_{ar\_K}}$, ($Q_{ar\_I}$, girl, holds), UsedFor)'' as the program for the question ``what is the object that the girl is holding used for'', which should be answered by first locating the ``hotdog'', then answer its functionality.
The program layouts are used as the groundtruth inference process and can be used for deep inspection of VQA models.

\subsection{Template-based generation.} 
There are a total of $7$ question composition types and corresponding templates that include different reasoning steps and various knowledge involvements, as depicted in Table~\ref{tab:templates}. Given the sampled triplets, query type, and composed program, we can determine its question type \textit{Qtype} and use the corresponding templates to generate the question-answer pairs. For instance, given a triplet ``(man, holds, umbrella)'' and \textit{Qtype 1}, we use the template ``\textit{Que}: what is $<$A$>$ $<$R$>$ \textit{Ans}: $<$B$>$'' to generate the question ``what is the man holding'' and the answer ``umbrella''. The reasoning step of a question can also be determined by the number of the selected triplets. For example, ``(a, r, b)'' corresponds to one-step question, and ``$(a, r_1, b)-(b, r_2, c)$'' is used to generate the two-step question. A question is considered as ``KB-related'' if one of its triplets is sampled from the knowledge base; otherwise, it is ``KB-not-related''. For example, the question above is a ``KB-not-related'' question because ``(man, holds, umbrella)'' can be detected in the image. However, ``What is the usage of the object that the man is holding?'', which is converted from ``(man, holds, umbrella)-(umbrella, used for, keep out rain)'', is ``KB-related'' because ``(umbrella, used for, keep out rain)'' only appears in the knowledge base.

We also add extra constraints to the randomly sampled process to avoid answer ambiguity and minimize the dataset biases. Specifically, we filter out invalid questions that have more than one answer or have no answer. We also restrict the answer frequency such that each answer cannot appear more than 100 times for each ``KB-related'' question types. To prevent the model from memorizing the question-knowledge pairs, a knowledge triplet cannot appear more than once for all questions of type: \textit{qtype 2}, \textit{qtype 3}, and \textit{qtype 5}.

% \begin{figure}[t]
% 	\centering
% 	\includegraphics[width=\columnwidth]{qlengths.png}
% 	\caption{Distributions of the question lengths in KRVQA.
% 	}
% 	\label{fig:qlengths}
% \end{figure}

% \begin{figure}[t]
% 	\centering
% 	\includegraphics[width=\columnwidth]{tans_nkb.png}
% 	\caption{Top-15 answer frequence of the KB-not-related questions.
% 	}
% 	\label{fig:ans_nkb}
% \end{figure}
\begin{figure}[t]
	\centering
	\begin{subfigure}[b]{\columnwidth}
         \centering
         \includegraphics[width=\textwidth]{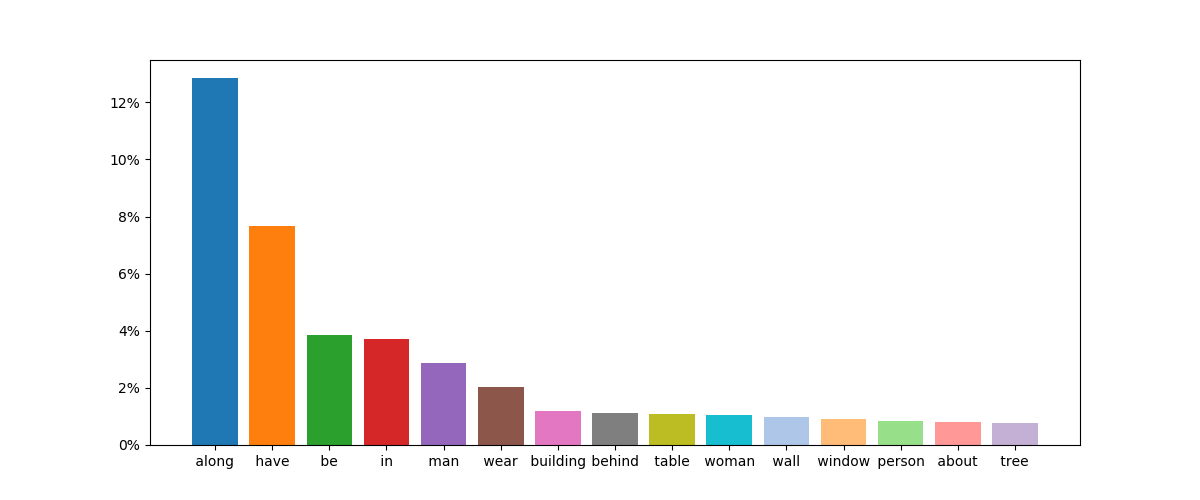}
         \caption{}
         \label{fig:ans_nkb}
     \end{subfigure}
     \vfill
     \begin{subfigure}[b]{\columnwidth}
         \centering
         \includegraphics[width=\textwidth]{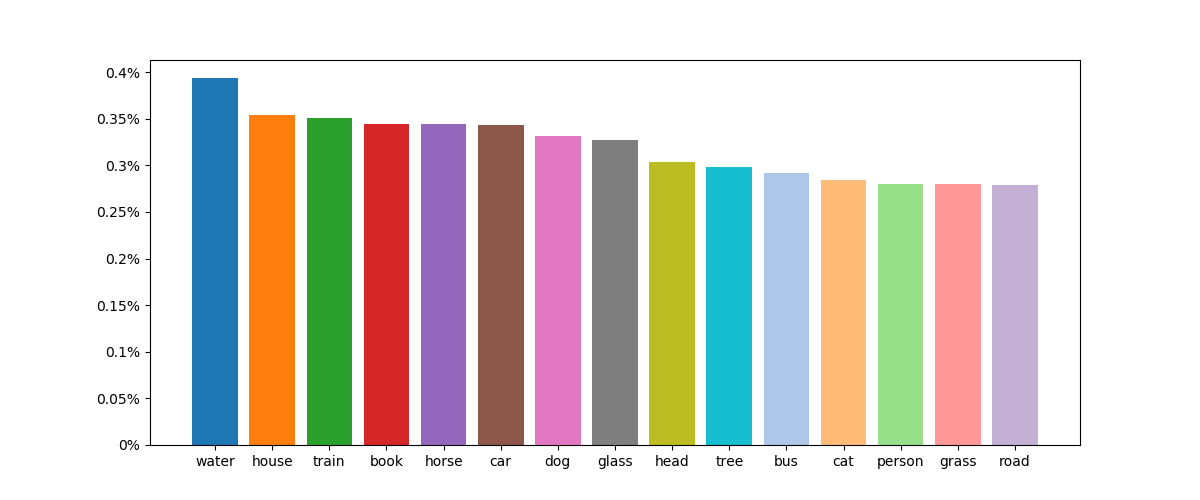}
         \caption{}
         \label{fig:ans_kb}
     \end{subfigure}
     \caption{The answer distribution of (a) KB-not-related questions and (b) KB-related questions respectively.}
\end{figure}

\section{Dataset Statistics}
Our KRVQA dataset contains 32,910 images and 157,201 QA pairs. They are split into the train, validation, and test set at the ratio of 60\%, 20\%, and 20\% respectively, as shown in Table~\ref{tab:data-all}. According to the reasoning steps, questions can be divided into {\bf one-step} and {\bf two-step} types of questions, which consist of 68,448 and 88,753 samples respectively. Based on the knowledge involvement, a question is divided into {\bf KB-related} and {\bf KB-not-related}, with 87,193 and 70,008 samples respectively. The lengths of the questions range from 4 to 24 and are 11.7 on average. Longer questions correspond to samples that require more reasoning steps.

\textbf{Answer distribution.} For the KB-not-related questions, the answer vocabulary size is 2,378 and shows a long-tailed distribution, as shown in Figure~\ref{fig:ans_nkb}. For the KB-related questions, the answer vocabulary size is 6,536, which is significantly greater than that of KB-not-related questions because of the large scale words in the knowledge base. A total of 97\% of the answers in the validation and test set can be found in the train split. As shown in Figure~\ref{fig:ans_kb}, the distribution of the KB-related answers is balanced as we limit the maximal presence count for each answer.

\textbf{Knowledge base.} Following~\cite{FVQA}, the knowledge base contains 193,449 knowledge triplets, 2,339 types of relations and 102,343 distinct entities. We restrict each triplet to present only once for \textit{Qtype 2}, \textit{Qtype 3}, and \textit{Qtype 5}, results in a total of 45,550 different knowledge triplets involve in the questions.

\begin{table}[t!]
	\centering
	\caption{Number of questions of different types in KRVQA.
	}
	\resizebox{\columnwidth}{!}{
		\begin{scriptsize}
			\begin{tabular}{|c|c|c|c|c|c|}
				\hline
				Step & qtype & Train & Val & Test & Total \\
				\hline
				\multirow{3}{*}{1} & 0 & 8,133 & 2,730 & 2,698 & 13,561\\
				& 1 & 7,996 & 2,660 & 2,734 & 13,390\\
				& 2 & 25,171 & 7,982 & 8,344 & 41,497\\
				\hline
				\multirow{4}{*}{2} &3 & 9,477 & 2,931 & 3,131 & 15,539\\
				& 4 & 16,498 & 5,564 & 5,505 & 27,567\\
				& 5 & 10,482 & 3,291 & 3,518 & 17,291\\
				& 6 & 17,058 & 5,518 & 5,780 & 28,356\\
				\hline
			\end{tabular}
		\end{scriptsize}
	}
	\label{tab:data-all}
\end{table}

\begin{table*}[t!]
	\centering
	\caption{The answering accuracy (\%) of different baseline and state-of-the-art VQA methods on the KRVQA dataset. Each column presents the accuracy of each question type.}
	\resizebox{\textwidth}{!}{
		\begin{tabular}{l|ccc|cccc|c|cccc|c}
			& \multicolumn{7}{c|}{KB-not-related} & \multicolumn{5}{c|}{KB-related} &\\
			& \multicolumn{3}{c|}{one-step} & \multicolumn{4}{c|}{two-step} & one-step & \multicolumn{4}{c|}{two-step} & \\
			Method & 0 & 1 & 2 & 3 & 4 & 5 & 6 & 2 & 3 & 4 & 5 & 6 & Overall\\
			\hline
			\hline
			Q-type & 36.19 & 2.78 & 8.21 & 33.18 & 35.97 & 3.66 & 8.06 & 0.09 & 0.00 & 0.18 & 0.06 & 0.33 & 8.12\\
			LSTM & 45.98 & 2.79 & 2.75 & 43.26 & 40.67 & 2.62 & 1.72 & 0.43 & 0.00 & 0.52 & 1.65 & 0.74 & 8.81 \\
			Program Predict & 58.86 & 50.98 & 59.17 & 54.71 & 57.31 & 54.17 & 57.64 & 65.16 & 33.95 & 71.64 & 63.05 & 76.53 & 61.62\\
			FiLM~\cite{film} & 52.42 & 21.35 & 18.50 & 45.23 & 42.36 & 21.32 & 15.44 & 6.27 & 5.48 & 4.37 & 4.41 & 7.19 & 16.89\\
			MFH~\cite{mfh} & 43.74 & 28.28 & 27.49 & 38.71 & 36.48 & 20.77 & 21.01 & 12.97 & 5.10 & 6.05 & 5.02 & 14.38 & 19.55\\
			UpDn~\cite{buattention} & {56.42} & {29.89} & {28.63} & {49.69} & {43.87} & {24.71} & {21.28} & 11.07 & 8.16 & 7.09 & 5.37 & 13.97 & 21.85\\
			MCAN~\cite{mcan} & 49.60 & 27.67 & 25.76 & 39.69 & 37.92 & 21.22 & 18.63 & 12.28 & 9.35 & 9.22 & 5.23 & 13.34 & 20.52\\
			\hline
			
		\end{tabular}
	}
	
	\label{tab:acc}
\end{table*}

\begin{table*}[t!]
	\centering
	\caption{The answering accuracy (\%) of different methods of incorporating knowledge embedding on the KRVQA dataset.}
	\resizebox{\textwidth}{!}{
		\begin{tabular}{l|ccc|cccc|c|cccc|c}
			& \multicolumn{7}{c|}{KB-not-related} & \multicolumn{5}{c|}{KB-related} &\\
			& \multicolumn{3}{c|}{one-step} & \multicolumn{4}{c|}{two-step} & one-step & \multicolumn{4}{c|}{two-step} & \\
			Method & 0 & 1 & 2 & 3 & 4 & 5 & 6 & 2 & 3 & 4 & 5 & 6 & Overall\\
			\hline
			Baseline(MCAN) & 49.60 & 27.67 & 25.76 & 39.69 & 37.92 & 21.22 & 18.63 & 12.28 & 9.35 & 9.22 & 5.23 & 13.34 & 20.52\\
			+ GT knowledge triplet & 50.55 & 27.18 & 24.90 & 41.46 & 37.29 & 23.05 & 20.83 & 90.90 & 75.83 & 12.67 & 63.66 & 16.51 & 42.08\\
			+ GT knowledge entry & 50.51 & 27.44 & 26.82 & 44.37 & 37.92 & 22.24 & 21.86 & 94.67 & 88.13 & 13.40 & 74.56 & 18.02 & 44.82\\
			\hline
			+ knowledge inference & 52.38 & 26.92 & 24.26 & 46.28 & 41.56 & 22.10 & 20.37 & 22.75 & 54.18 & 10.75 & 17.84 & 13.56 & 26.34\\
			+ knowledge retrieval & 51.32 & 27.14 & 25.69 & 41.23 & 38.86 & 23.25 & 21.15 & 13.59 & 9.84 & 9.24 & 5.51 & 13.89 & 21.30\\
			+ knowledge embedding extraction & 50.33 & 27.56 & 25.20 & 41.38 & 37.54 & 22.51 & 20.31 & 12.48 & 9.23 & 9.32 & 4.74 & 12.96 & 20.69\\
			\hline
			w/ Masked-pretrain & 54.58 & 26.88 & 25.01 & 46.82 & 42.06 & 23.66 & 20.95 & 15.11 & 10.46 & 9.52 & 5.12 & 13.84 & 22.23\\
			w/ Triplet-pretrain & 51.10 & 25.68 & 24.48 & 43.45 & 41.10 & 21.83 & 21.28 & 12.07 & 9.35 & 9.24 & 5.07 & 13.66 & 20.86\\
		\end{tabular}
	}
	\label{tab:abl}
\end{table*}

\begin{figure*}[t!]
	\includegraphics[width=\textwidth]{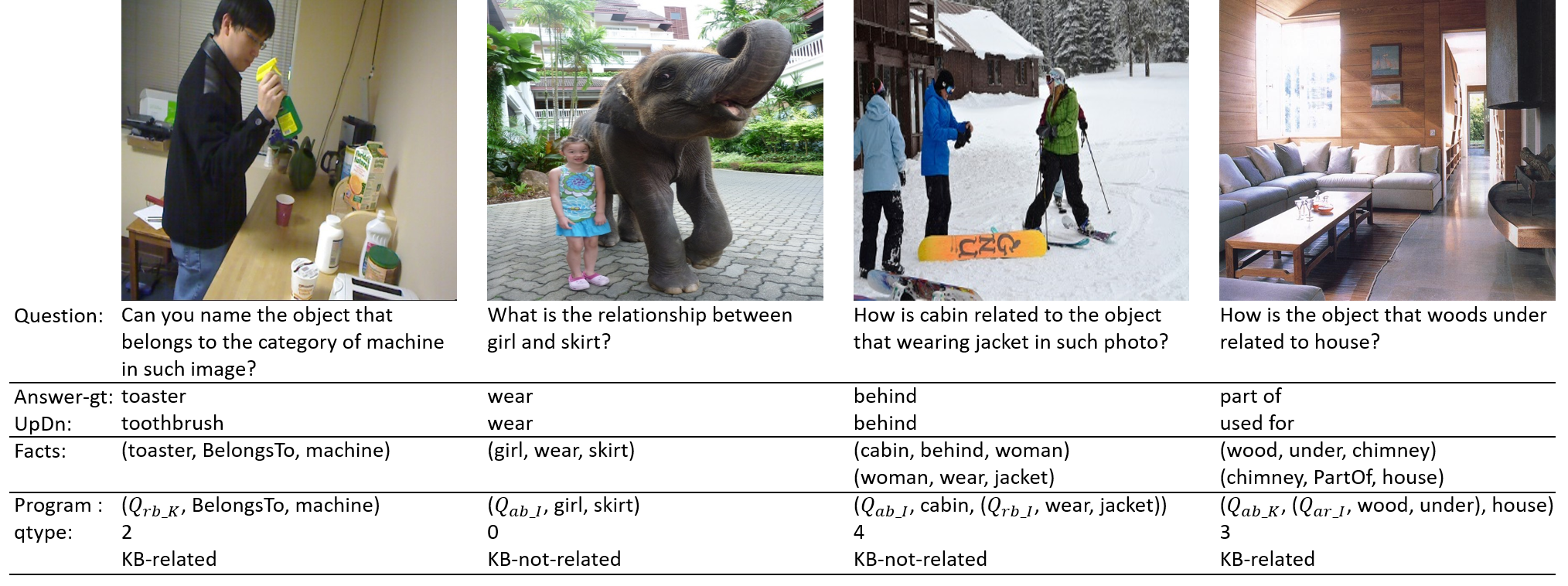}
	\caption{Examples of the KRVQA and the results predicted by the bottom-up attention baseline method. The ``Answer-gt'' refers to the ground-truth answer and the ``Fact'' refers to the ground-truth supporting facts.
	}
	\label{fig:dataviz}
\end{figure*}

\begin{figure}[t]
	\includegraphics[width=\columnwidth]{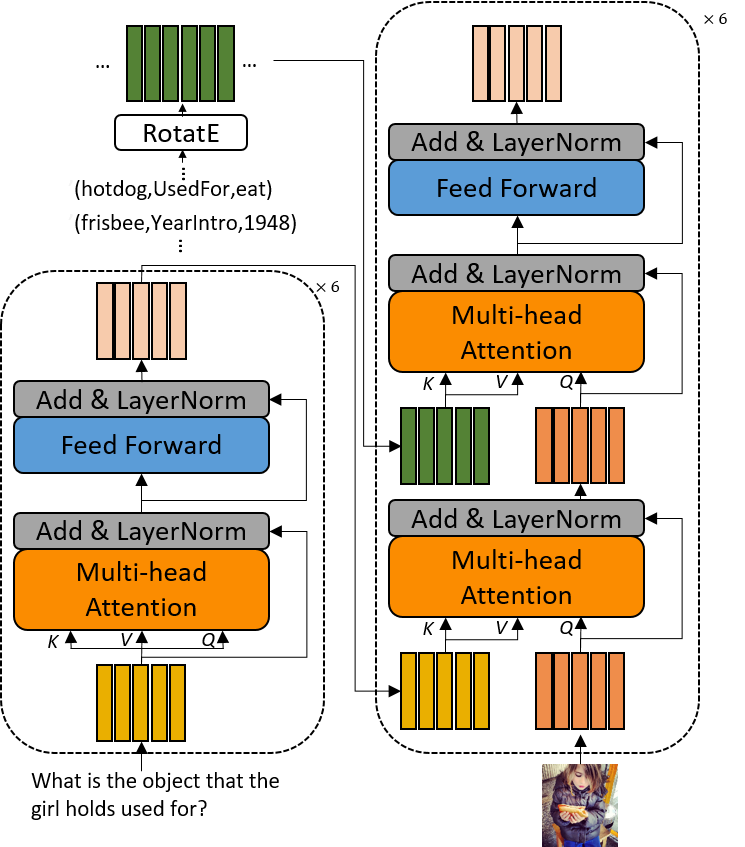}
	\caption{The overview architecture of MCAN with the knowledge guided-attention block.}
	\label{fig:arc_mcan}
\end{figure}

\section{Experiments}
We first evaluate the performance of multiple state-of-the-art VQA models on our proposed KRVQA dataset. Then, we perform extensive experiments to inspect the properties of our proposed KRVQA dataset and evaluate a wide range of knowledge embedding methods, including knowledge graph embedding and question encoder pretraining.

\subsection{Answering Accuracy of baseline and state-of-the-art VQA methods}
We evaluate the following baseline and state-of-the-art methods on the KRVQA dataset.

\textbf{Q-type}. For each question, we use the most frequent training answer of its question type as the output answer. We use this baseline to inspect the question-answer bias.

\textbf{LSTM}. We first embed the question word into a $1024$-dimensional vector, then use a bi-directional LSTM with $2048$ hidden-size for each direction to encode the whole questions. The hidden vector of the last word is used as question embedding and is fed into a two-layer MLP that has $4096$ hidden-size to predict the final answer.

\textbf{Program Prediction}. Given the question type and program that is used to generate the question-answer pair, one can retrieve the correct answer from the scene graph and knowledge base. We thus train a model to predict the program and question type with full supervision as a baseline that can access to all possible information. Similar to the ``LSTM'' baseline, this model embeds the word and question with a $1024$-dimensional lookup table and a $2048$-dimensional bi-directional GRU. The question embedding is used to predict the question type and fed into a $4096$-dimensional GRU decoder to predict the sequential program. A sample is considered correct if both the predicted question type and program are correct.

\textbf{Bottom-up attention}. We evaluate the 1st-place method~\cite{tips} of the 2017 VQA challenge. This method uses a GRU to encode the questions and extracts the image feature using a Fast-RCNN trained on the Visual Genome dataset. Then it performs multimodel bilinear pooling~\cite{MLB} and attention mechanism on the image object feature and question embedding. The attended feature is passed through an MLP to predict the answer. We use the top-$36$ object features that have the highest object scores as our image input.

\textbf{MFH}. The Multi-modal Factorized High-order pooling(MFH)~\cite{mfh} extracts the image object features similar to the bottom-up attention and encodes the question with an LSTM. It then performs attention and question and image separately. The attention features are fused using stacks of bilinear fusion blocks~\cite{mfh}. The answer is predicted by passing the fused feature through an MLP.

\textbf{FiLM}. The Feature-wise Linear Modulation(FiLM)~\cite{film} embeds the question with GRU and embeds the image with multiple convolutional blocks. In each block, the question embedding is transformed into two vectors, and is multiplied and added with the normalized feature maps respectively. It is a simple but efficient model that has achieved high accuracy on the CLEVR~\cite{clevr} dataset, which requires compositional reasoning ability.

\textbf{MCAN}. The Modular Co-Attention Network(MCAN)~\cite{mcan} consists of $6$ self-attention blocks to encode the question words and $6$ guided-attention blocks to encode the image objects. The object features are fused with the question words at each guided-attention block and both the question and image outputs are used to predict the answer. The MCAN currently achieves the highest accuracy on the VQAv2 dataset without large amounts of pretraining data and has similar architecture blocks with other state-of-the-art works pretrained on various vision-language tasks.

The answering accuracy of each method on different question types is shown in Table~\ref{tab:acc}. The two baselines ``Q-type'' and ``GRU'' predict the answer based on the question solely and achieve the accuracies of $8.12\%$ and $8.81\%$ respectively. These two Q-only baselines perform much worse than other methods, suggesting the image context is necessary to answer the questions in our proposed dataset. The state-of-the-art VQA models also have low accuracies compared to their performances on previous VQA datasets (e.g., VQAv2, CLEVR, etc.).
The FiLM improves the accuracy by a large margin compared with the previous two baselines as it uses the image context. It is interesting that this bottom-up method performs better than other MFH and MCAN, suggesting that a complex method might be overfitting on our training set.
All the methods perform worse on two-step questions than on one-step questions. These results show that multi-hop reasoning is extremely challenging, and also imply that the proposed dataset requires incorporating knowledge while reasoning.

\begin{figure*}[htbp]
 \centering
 \includegraphics[width=0.90\textwidth]{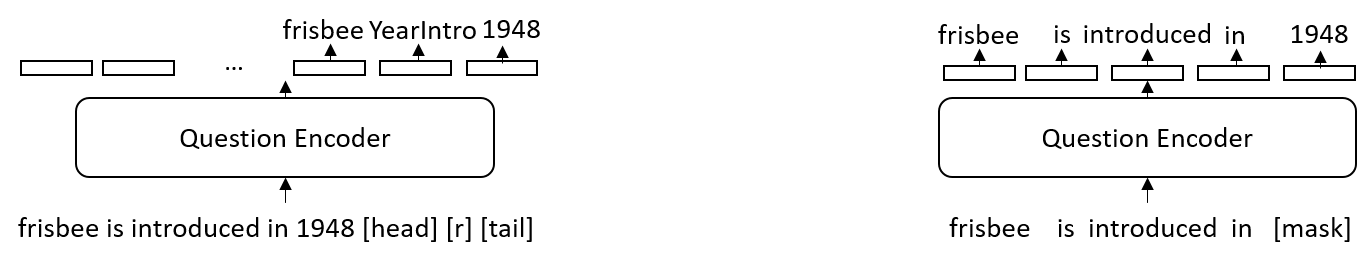}
 \caption{The illustration of the two pretraining task settings. On the left, the knowledge surface text is appended with three extra tokens and the encoder is trained to predict the corresponding knowledge triplets. The question encoder on the right is trained to predict the masked token in the knowledge surface text.}
 \label{fig:pretrain}
\end{figure*}

\begin{figure}[htbp]
	\includegraphics[width=\columnwidth]{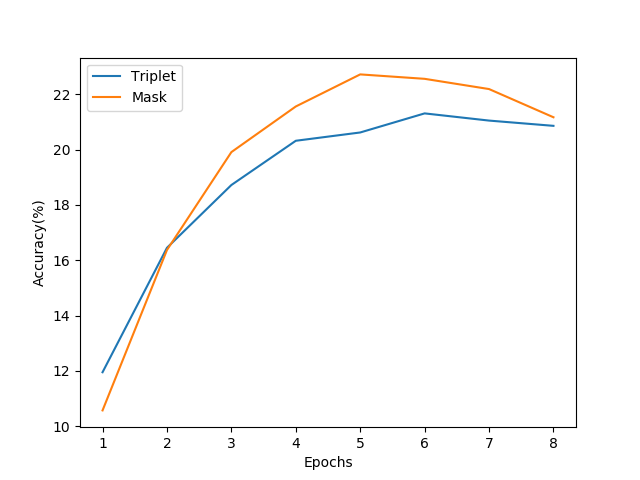}
	\caption{The test answering accuracy (\%) of the MCAN on KRVQA with respect to the training epochs. The blue line represents the MCAN accuracy with question encoder pretrained to predict knowledge triplets. And the orange line represents that the question encoder is pretrained on the masked token prediction task.}
	\label{fig:acc_curve}
\end{figure}

\subsection{Knowledge graph embedding}
In this section, we evaluate several baseline methods that incorporate the embedding knowledge graph into the VQA model, i.e., MCAN. We use MCAN as baseline architecture because it achieves state-of-the-art results on the VQAv2 dataset and shares the similar self-attention and guided-attention modules with other models that pretrained on large amounts of extra data.
Specifically, we first train the RotatE~\cite{rotate} on the whole knowledge base to extract the entities and relation features. Then, we modify the guided-attention block to fuse the knowledge embedding vectors with the image object feature in a similar way as fusing the question words. We first give a simple description of the RotatE method and our modified MCAN, then present the accuracy results of different settings.

\textbf{Knowledge graph embedding and RotatE}
Given a knowledge graph consists of triplets (head entity, relation, and tail entity), the knowledge graph embedding method focuses on learning the representations of entities and relations for predicting the missing links. For example, TransE~\cite{transe} is trained to encode a triplet as vectors $(h,r,t)$ such that $h+r=t$. Thus, given two entry of a knowledge triplet, one can obtain the missing entry by performing corresponding operations on the embedding vector, such as $t=h+r$, $r=h-t$, etc.

The RotatE maps the embedding vectors to complex space and treats the relation as a rotation from head entity to tail entity. Thus, the entity has the from $h+ih_i$ and relation has $\cos r + i\sin r_i$. During training, it minimizes the distance between the rotated head entity and the tail entity, which is $|(h+ih_i)(\cos r + i\sin r_i)-(t+it_i)|^2$. Thus, we use the following equations to obtain the tail entity:
\begin{align}
\begin{split}\label{eq:kemb_t}
    t &= h * \cos r - h_i * \sin r_i \\
    t_i &= h * \sin r_i + h_i * \cos r
\end{split} \\
\begin{split}\label{eq:kemb_r}
    \cos r &= (h*t + h_i*t_i) / (h^2+h_i^2) \\
    \sin r_i &= h*t_i - h_i*t / (h^2+h_i^2)
\end{split} \\
\begin{split}\label{eq:kemb_h}
    h &= t * \cos r + t_i * \sin r_i \\
    h_i &= t_i * \cos r - t * \sin r_i
\end{split}
\end{align} 

%  test MR: 1.35 | MRR: 0.966 | H@1: 0.945 | H@3: 0.986 | H@10: 0.998
We train the RotatE on the provided knowledge graph and use $500$-dimensional vectors to represent both the real part and the imaginary part. After $50$ epochs, the RotatE can predict the tail entity with a mean rank $1.35$ and top-$1$ accuracy $94.5\%$. We extract the entity and relation embedding and concatenate the real and imaginary parts to represent each entity and relation with a $1000$-dimensional vector.  

\textbf{Knowledge guided-attention block}
We modify the guided-attention block of the original MCAN to incorporate the extracted knowledge embedding. As shown in Figure~\ref{fig:arc_mcan}, the guided-attention block first performs multi-head attention on image and question respectively. Then we add another multi-head attention whose key and value are based on knowledge embedding and the query is conditioned on the image. This layer can attend the candidate entity and relation embedding from the knowledge graph, and fuse them into the image representation. Other parts the MCAN remains the same for final answer prediction.

\textbf{MCAN with groundtruth knowledge} We first evaluate a baseline setting that the supporting knowledge is provided. As shown in Table~\ref{tab:abl}, the knowledge-guided MCAN achieves $42.08\%$ and $44.82\%$ if the supporting triplet or entry is provided and the model with supporting entry performs slightly better. This model achieves similar results with the MCAN on KB-not-related question and KB-related question type 4 and 6. The answers to these questions are from the image, thus the knowledge embedding can't provide much information. However, the models improve dramatically on question types 2, 3, and 5 by achieving the accuracy higher than $70\%$. These results demonstrate the effeteness of the knowledge embedding vector.

\textbf{MCAN with knowledge inference} We then evaluate the model with supporting entry obtained with Equation~\ref{eq:kemb_t}-\ref{eq:kemb_h}.
Instead of providing the complete supporting triplet, we provide the model with two entries and then calculate the supporting entry with the above equations. The model achieves $26.34\%$ overall accuracy. Its performance drops a lot on KB-related question type 2 and 5, but drops relatively small on question type 3. Since the answers to question type 2 and 5 are from the image, where the answers are also related to multiple knowledge triplets according to our dataset constraint. For example, the triplets ``(car, faster, bicycle)'' and ``(plane, faster, bicycle)'' are both valid for the query ``($Q_{rb\_K}$, faster, bicycle)'', but only the ``car'' is present in the image. Thus, a model can't simply obtain the knowledge embedding to predict the answer without image context, which leads to the large performance drop on question type 2 and 5 only. This demonstrates the importance of correctly perceiving the image context and associating them with knowledge.

\textbf{MCAN with knowledge embedding extraction} We also train a model to extract the knowledge embedding from the question and use the extracted and inferred embedding with Equation~\ref{eq:kemb_t}-\ref{eq:kemb_h} as the guided-attention block input. Specifically, each knowledge triplet is associated with a surface text that describes this knowledge in natural language. We append the surface text with $3$ tokens ``[head]'', ``[r]'' and ``[tail]'', and encode the extended surface text with an encoder. The encoder has the same architecture with the question encoder in MCAN, and is trained to minimize the L2 distance between the embedding of the three tokens and the RotatE knowledge embedding. It achieves a train loss of $0.6$ and is used to extract the knowledge triplets embedding from questions. The extracted three entries and the three entries predicted with Equation~\ref{eq:kemb_t}-\ref{eq:kemb_h} are attended and fused by the image via the guided-attention block. We also update its parameters with the rest of the MCAN during VQA training.

As shown in Table~\ref{tab:abl}, this model achieves $20.69\%$ overall accuracy, which is almost the same with the baseline method and drops $5.65\%$ compared with the ``+ knowledge inference''. This illustrates that the current black-box model can't discover the sparse knowledge through answer supervision only and validates that our proposed dataset can't be simply fit by deep models.

\textbf{MCAN with related fact} We retrieve the knowledge triplets from the knowledge base by matching the question tokens with the triplet entries. Specifically, we first extract the noun words from the question and sorts them according to their term frequency-inverse document frequency (TF-IDF). Then we match the words with the entities in the knowledge base and retrieve all triplets that are related to the matched entity. We fed the top-16 triplets to MCAN which performs attention on these $16$ triplets. Table~\ref{tab:abl} shows that its result only improves $0.78\%$, compared with the baseline MCAN. This implies that only a small fraction of samples can retrieve the supporting knowledge, showing the difficulty of handling a large knowledge base.

\subsection{Knowledge text pretraining}
Most recent works~\cite{comet,kbert} have shown that the language model can encode the knowledge to a certain extend by training on a large corpus of text. Similar to these works, we pretrain the question encoder on knowledge text to encode the knowledge implicitly. Specifically, as each knowledge triplet has a corresponding surface text, we pretrain the self-attention question encoder of the MCAN to use the surface text as input, then either predict the corresponding triplets or predict the masked text tokens.

\textbf{Triplet classification} We first pretrain the question encoder to predict the knowledge triplet. As shown in Figure~\ref{fig:pretrain}, we add three extra tokens to the end of the surface text, then fed the embedding of these three tokens to a fully-connected layer to predict the head entity, relation, and tail entity respectively. After training 30 epochs, the encoder predicts the head, relation, and tail with the accuracy of $99.64\%$, $99.90\%$ and $98.82\%$ respectively. We initialize the question encoder in MCAN with the pretrained parameters and it achieves $20.86\%$, as shown in Table~\ref{tab:abl}.

\textbf{Masked token classification} As shown in Figure~\ref{fig:pretrain}, we randomly mask $1$ or $2$ tokens in the surface text then train the question decoder to predict the masked tokens. With 240 epochs training, the question encoder achieves token classification accuracy of $97.16\%$. And the MCAN initialized with the pretrained parameters has the answer accuracy of $22.23\%$, as shown in Table~\ref{tab:abl}.

We further investigate the test accuracy on different training epochs. As can be seen in Figure~\ref{fig:acc_curve}, the two pretrained models achieve their best results on epoch $5$ and $6$ respectively. Then, their test accuracies drop as the train loss decrease. This demonstrate that the pretrained question encoder has encoded useful information, but the deep embedding model overfits the training set later and can't handle the unused triplets very well.

\section{Conclusion}
In this paper, we presented our Knowledge-Routed Visual Question Reasoning (KRVQA) dataset to evaluate a VQA method's ability on knowledge visual question reasoning and reveal the drawback of current deep embedding models. Specifically, we leverage the scene graph and knowledge base to retrieve the supporting facts and use the tightly controlled programs to generate the questions that require multi-step reasoning on external knowledge. The extensive and comprehensive experiments have justified that the current VQA models can't achieve high performances by exploiting shortcuts, due to their difficulties in handling large scale of knowledge base and retrieving the supporting one for answer prediction. In future work, we will attempt to develop novel yet powerful VQA models and evaluate its performance on this proposed dataset.

\section*{Acknowledgment}
This work was supported in part by the State Key Development Program under Grant 2018YFC0830103, in part by the National Natural Science Foundation of China under Grant No. U181146, 61836012 and 62006255, and in part by Guangdong Natural Science Foundation under Grant No. 61976233, 2017A030312006 and 2017A03031335, and DARPA XAI project N66001-17-2-4029.

\section*{Appendix}
The supplementary material presents more randomly selected questions and its experimental results from HVQR. Each question is annotated with its answer, qtype, predicted supporting facts and the reasoning process of KM-net according to the predicted query layout.

\clearpage
\begin{figure}[t]
    \includegraphics[width=\textwidth]{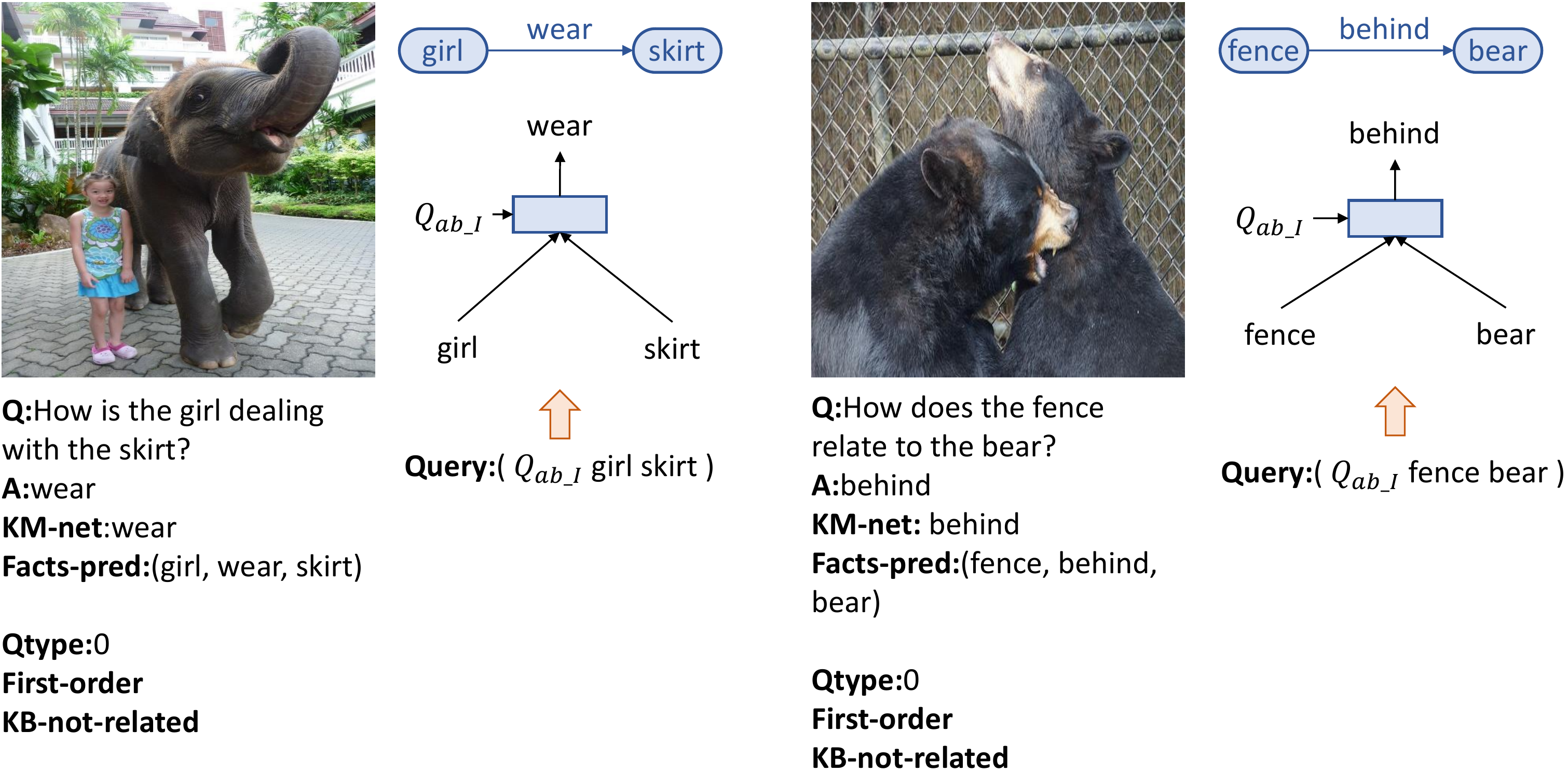}
    \label{fig:eg1}
    % \vspace{-5mm}
\end{figure}
\begin{figure}[!t]
    \includegraphics[width=\textwidth]{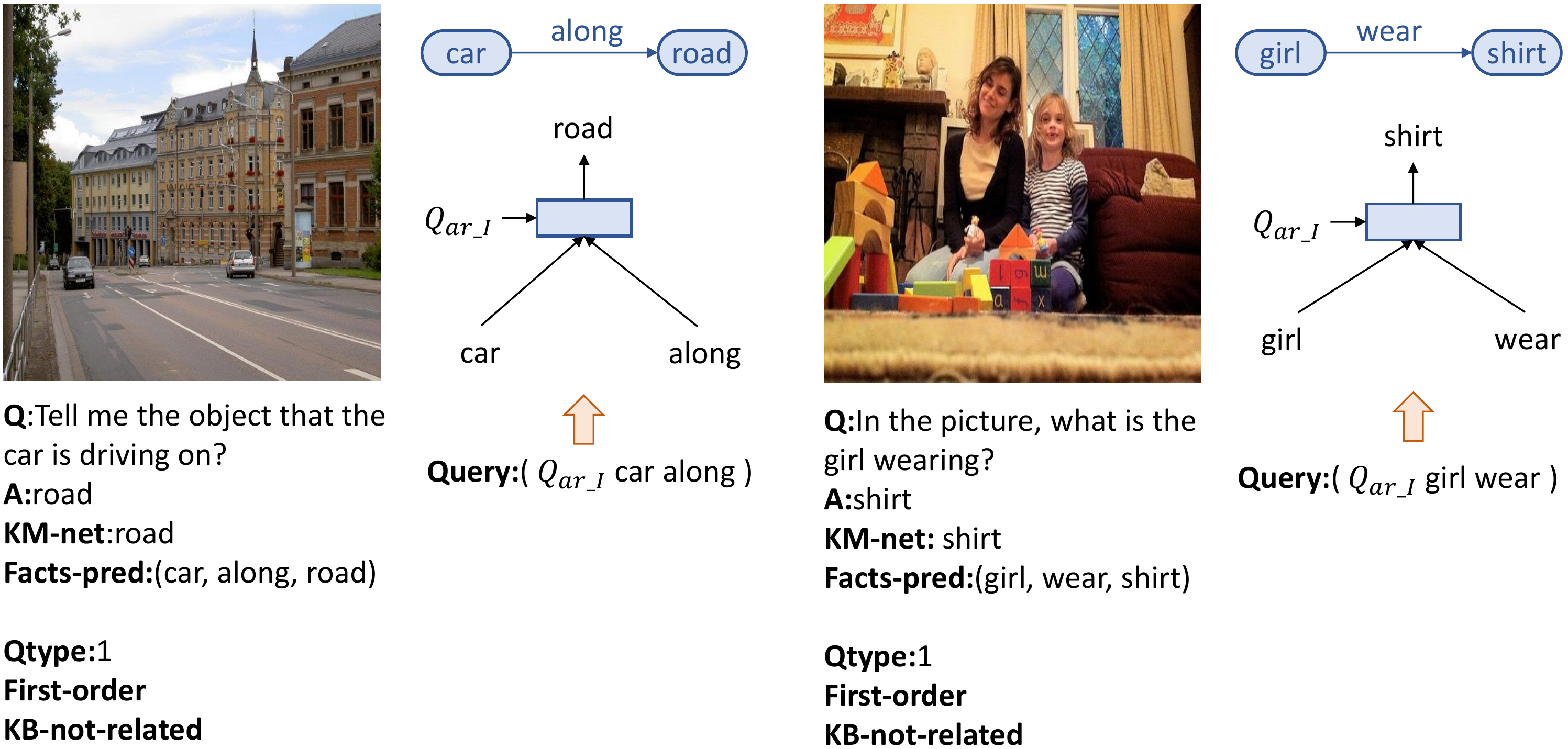}
    \label{fig:eg1}
    % \vspace{-5mm}
\end{figure}
%-------------------------------------------------------------------------
\clearpage
\begin{figure}[!t]
    \includegraphics[width=\textwidth]{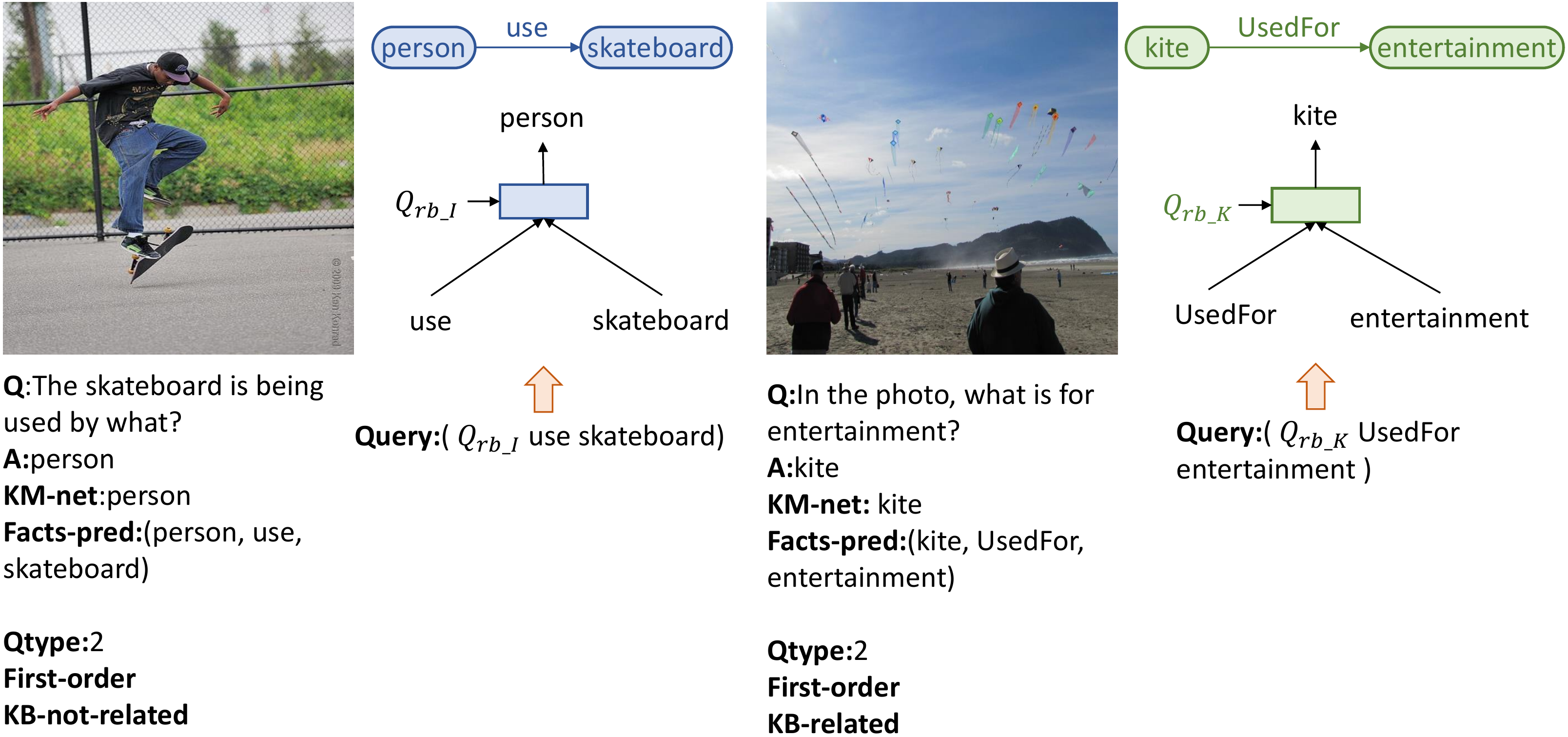}
    \label{fig:eg1}
    % \vspace{-5mm}
\end{figure}

\begin{figure}[!t]
    \includegraphics[width=\textwidth]{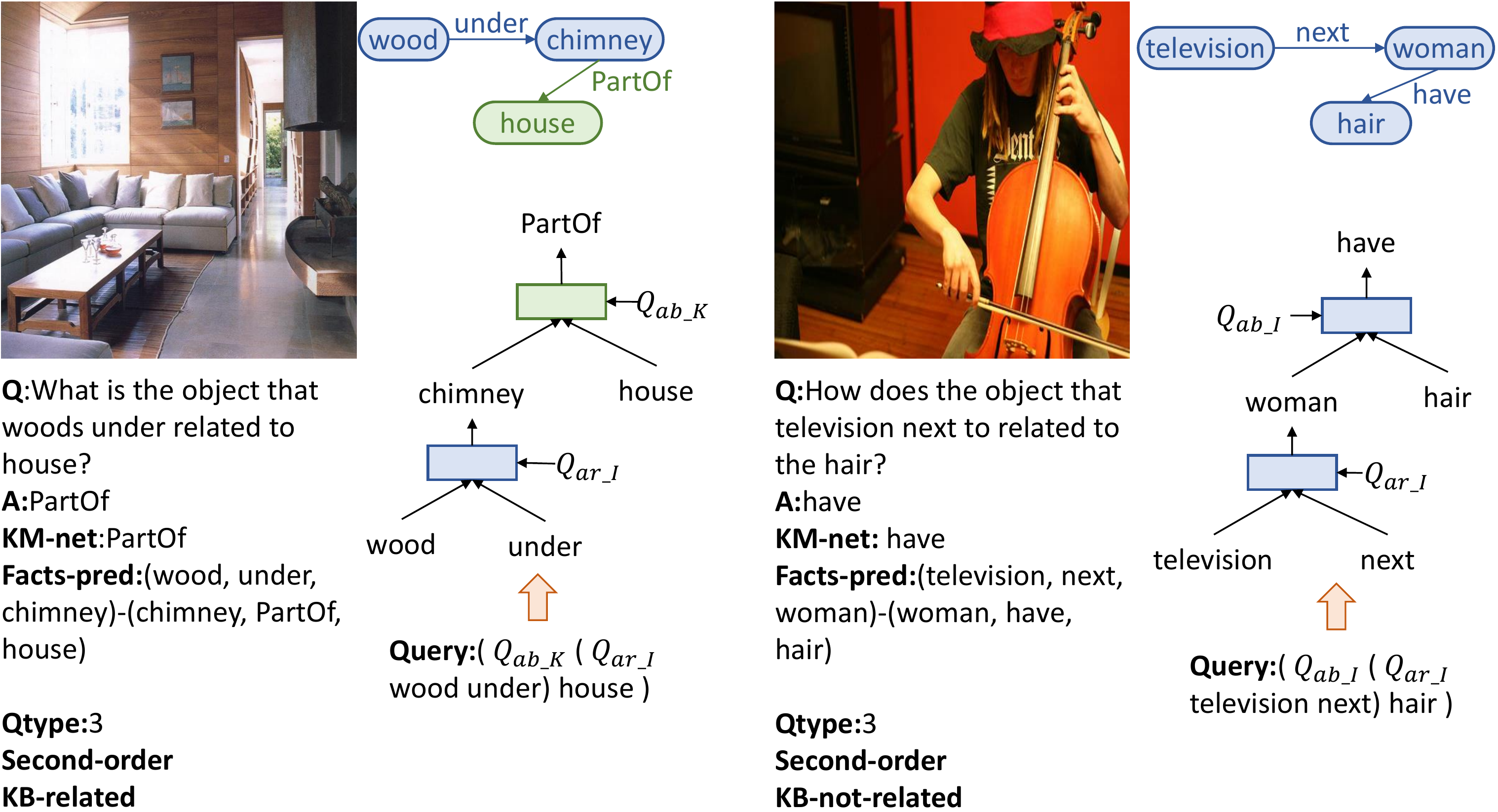}
    \label{fig:eg1}
    % \vspace{-5mm}
\end{figure}
\clearpage
\begin{figure}[!t]
    \includegraphics[width=\textwidth]{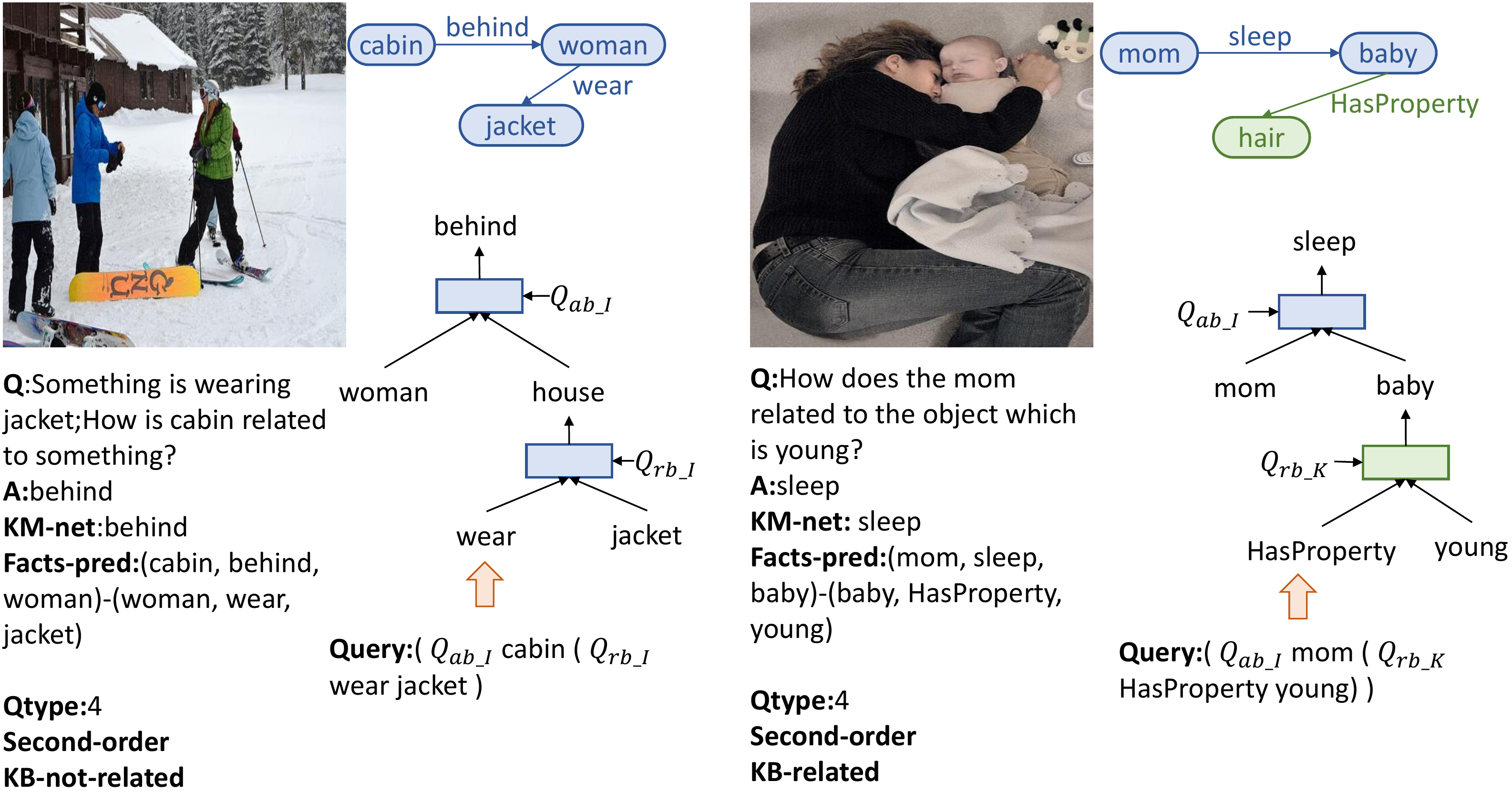}
    \label{fig:eg1}
    % \vspace{-5mm}
\end{figure}

\begin{figure}[!t]
    \includegraphics[width=\textwidth]{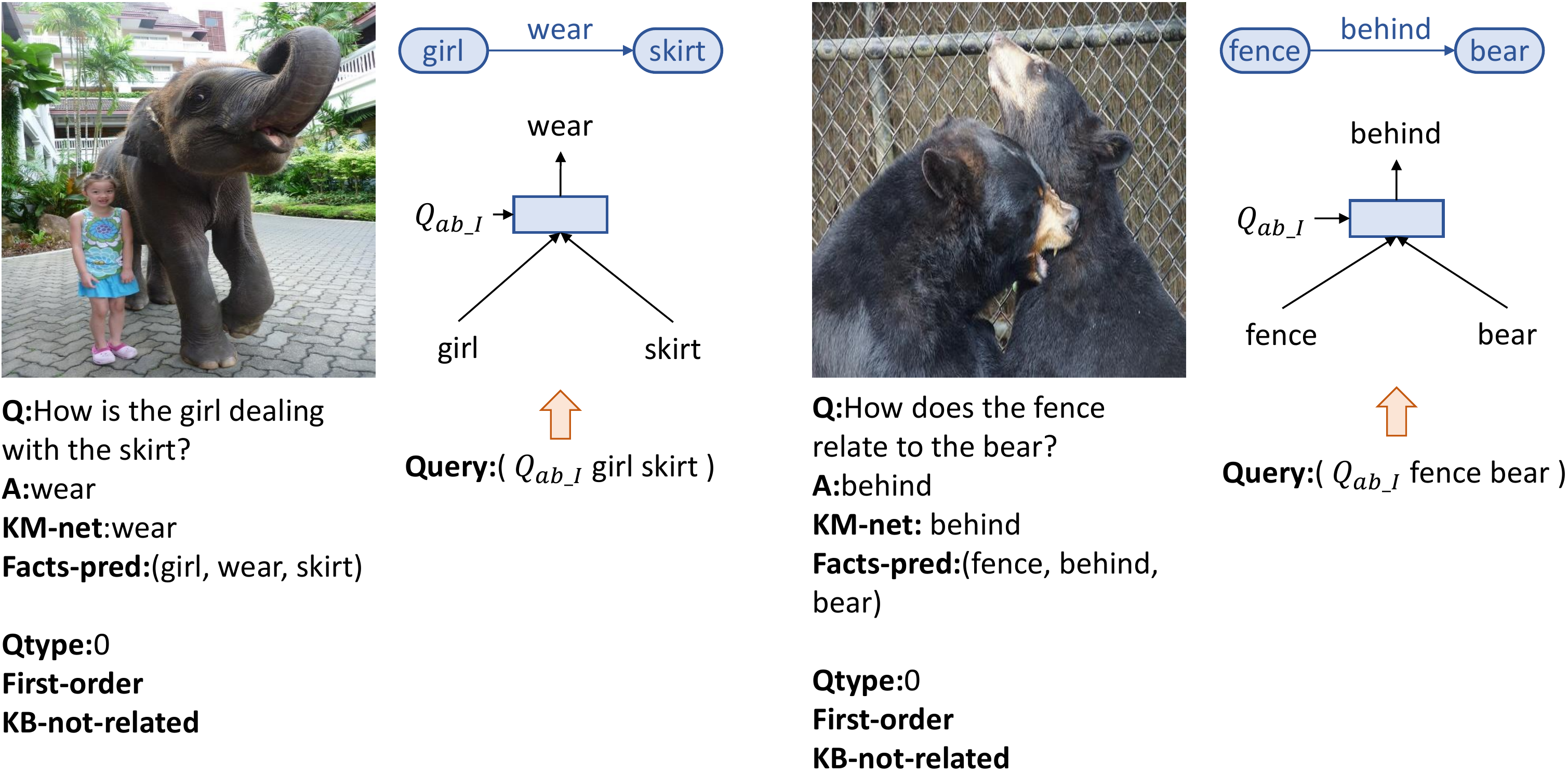}
    \label{fig:eg1}
    % \vspace{-5mm}
\end{figure}
\clearpage

% Can use something like this to put references on a page
% by themselves when using endfloat and the captionsoff option.
\ifCLASSOPTIONcaptionsoff
\newpage
\fi
\bibliographystyle{IEEEtran}
\bibliography{egbib}

% Generated by IEEEtran.bst, version: 1.14 (2015/08/26)
\begin{thebibliography}{10}
\providecommand{\url}[1]{#1}
\csname url@samestyle\endcsname
\providecommand{\newblock}{\relax}
\providecommand{\bibinfo}[2]{#2}
\providecommand{\BIBentrySTDinterwordspacing}{\spaceskip=0pt\relax}
\providecommand{\BIBentryALTinterwordstretchfactor}{4}
\providecommand{\BIBentryALTinterwordspacing}{\spaceskip=\fontdimen2\font plus
\BIBentryALTinterwordstretchfactor\fontdimen3\font minus
  \fontdimen4\font\relax}
\providecommand{\BIBforeignlanguage}[2]{{%
\expandafter\ifx\csname l@#1\endcsname\relax
\typeout{** WARNING: IEEEtran.bst: No hyphenation pattern has been}%
\typeout{** loaded for the language `#1'. Using the pattern for}%
\typeout{** the default language instead.}%
\else
\language=\csname l@#1\endcsname
\fi
#2}}
\providecommand{\BIBdecl}{\relax}
\BIBdecl

\bibitem{VQA}
S.~Antol, A.~Agrawal, J.~Lu, M.~Mitchell, D.~Batra, C.~L. Zitnick, and
  D.~Parikh, ``{VQA}: {V}isual {Q}uestion {A}nswering,'' in \emph{ICCV}, 2015.

\bibitem{balanced_vqa_v2}
Y.~Goyal, T.~Khot, D.~Summers{-}Stay, D.~Batra, and D.~Parikh, ``Making the {V}
  in {VQA} matter: Elevating the role of image understanding in {V}isual
  {Q}uestion {A}nswering,'' in \emph{CVPR}, 2017.

\bibitem{clevr}
J.~Johnson, B.~Hariharan, L.~van~der Maaten, L.~Fei-Fei, C.~L. Zitnick, and
  R.~Girshick, ``{CLEVR}: A diagnostic dataset for compositional language and
  elementary visual reasoning,'' in \emph{{CVPR}}, 2017.

\bibitem{GQA}
D.~A. Hudson and C.~D. Manning, ``Gqa: A new dataset for real-world visual
  reasoning and compositional question answering,'' \emph{CVPR}, 2019.

\bibitem{MLB}
J.-H. Kim, K.~W. On, J.~Kim, J.~Ha, and B.-T. Zhang, ``Hadamard product for
  low-rank bilinear pooling,'' in \emph{ICLR}, 2017.

\bibitem{film}
E.~Perez, F.~Strub, H.~de~Vries, V.~Dumoulin, and A.~C. Courville, ``Film:
  Visual reasoning with a general conditioning layer,'' in \emph{AAAI}, 2018.

\bibitem{mcan}
Z.~Yu, J.~Yu, Y.~Cui, D.~Tao, and Q.~Tian, ``Deep modular co-attention networks
  for visual question answering,'' in \emph{CVPR}, 2019.

\bibitem{lxmert}
H.~Tan and M.~Bansal, ``{LXMERT}: Learning cross-modality encoder
  representations from transformers,'' in \emph{EMNLP}, 2019.

\bibitem{VG}
R.~Krishna, Y.~Zhu, O.~Groth, J.~Johnson, K.~Hata, J.~Kravitz, S.~Chen,
  Y.~Kalantidis, L.-J. Li, D.~A. Shamma, M.~S. Bernstein, and L.~Fei-Fei,
  ``Visual genome: Connecting language and vision using crowdsourced dense
  image annotations,'' \emph{IJCV}, vol. 123, no.~1, p. 32–73, May 2017.

\bibitem{FVQA}
P.~Wang, Q.~Wu, C.~Shen, A.~Dick, and A.~{van den Hengel}, ``{FVQA}: Fact-based
  visual question answering,'' \emph{IEEE TPAMI}, 2017.

\bibitem{okvqa}
K.~Marino, M.~Rastegari, A.~Farhadi, and R.~Mottaghi, ``Ok-vqa: A visual
  question answering benchmark requiring external knowledge,'' in \emph{CVPR},
  2019.

\bibitem{vcr}
R.~Zellers, Y.~Bisk, A.~Farhadi, and Y.~Choi, ``From recognition to cognition:
  Visual commonsense reasoning,'' in \emph{CVPR}, 2019.

\bibitem{DBLP:journals/corr/IlievskiYF16}
I.~Ilievski, S.~Yan, and J.~Feng, ``A focused dynamic attention model for
  visual question answering,'' \emph{CoRR}, vol. abs/1604.01485, 2016.

\bibitem{Xu2016}
H.~Xu and K.~Saenko, ``Ask, attend and answer: Exploring question-guided
  spatial attention for visual question answering,'' in \emph{ECCV}, 2016.

\bibitem{shih2016att}
K.~J. Shih, S.~Singh, and D.~Hoiem, ``Where to look: Focus regions for visual
  question answering,'' in \emph{CVPR}, 2016.

\bibitem{Visual7W}
Y.~Zhu, O.~Groth, M.~S. Bernstein, and L.~Fei{-}Fei, ``Visual7w: Grounded
  question answering in images,'' in \emph{CVPR}, 2016.

\bibitem{san}
Z.~Yang, X.~He, J.~Gao, L.~Deng, and A.~J. Smola, ``Stacked attention networks
  for image question answering,'' in \emph{CVPR}, 2016.

\bibitem{HiCoAtt}
J.~Lu, J.~Yang, D.~Batra, and D.~Parikh, ``Hierarchical question-image
  co-attention for visual question answering,'' in \emph{NIPS}, 2016.

\bibitem{MCB}
A.~Fukui, D.~H. Park, D.~Yang, A.~Rohrbach, T.~Darrell, and M.~Rohrbach,
  ``Multimodal compact bilinear pooling for visual question answering and
  visual grounding,'' in \emph{EMNLP}, 2016.

\bibitem{MFB}
Z.~Yu, J.~Yu, J.~Fan, and D.~Tao, ``Multi-modal factorized bilinear pooling
  with co-attention learning for visual question answering,'' in \emph{ICCV},
  2017.

\bibitem{MUTAN}
H.~Ben-younes, R.~Cadene, M.~Cord, and N.~Thome, ``Mutan: Multimodal tucker
  fusion for visual question answering,'' in \emph{ICCV}, 2017.

\bibitem{Anderson2017up-down}
P.~Anderson, X.~He, C.~Buehler, D.~Teney, M.~Johnson, S.~Gould, and L.~Zhang,
  ``Bottom-up and top-down attention for image captioning and visual question
  answering,'' in \emph{CVPR}, 2018.

\bibitem{vqav2winner}
D.~Teney, P.~Anderson, X.~He, and A.~van~den Hengel, ``Tips and tricks for
  visual question answering: Learnings from the 2017 challenge,'' in
  \emph{CVPR}, 2018.

\bibitem{vl-bert}
W.~Su, X.~Zhu, Y.~Cao, B.~Li, L.~Lu, F.~Wei, and J.~Dai, ``Vl-bert:
  Pre-training of generic visual-linguistic representations,'' in \emph{ICLR},
  2020.

\bibitem{comet}
A.~Bosselut, H.~Rashkin, M.~Sap, C.~Malaviya, A.~Celikyilmaz, and Y.~Choi,
  ``Comet: Commonsense transformers for automatic knowledge graph
  construction,'' in \emph{ACL}, 2019.

\bibitem{kbert}
W.~Liu, P.~Zhou, Z.~Zhao, Z.~Wang, Q.~Ju, H.~Deng, and P.~Wang, ``{K-BERT}:
  Enabling language representation with knowledge graph,'' in \emph{AAAI},
  2020.

\bibitem{nmn}
J.~Andreas, M.~Rohrbach, T.~Darrell, and D.~Klein, ``Neural module networks,''
  in \emph{CVPR}, 2016.

\bibitem{lnmn}
------, ``Learning to compose neural networks for question answering,'' in
  \emph{NAACL}, 2016.

\bibitem{e2emn}
R.~Hu, J.~Andreas, M.~Rohrbach, T.~Darrell, and K.~Saenko, ``Learning to
  reason: End-to-end module networks for visual question answering,'' in
  \emph{ICCV}, 2017.

\bibitem{norcliffebrown2018learning}
W.~Norcliffe-Brown, E.~Vafeias, and S.~Parisot, ``Learning conditioned graph
  structures for interpretable visual question answering,'' in \emph{NIPS},
  2018.

\bibitem{NeuralSymbolic}
K.~Yi, J.~Wu, C.~Gan, A.~Torralba, P.~Kohli, and J.~B. Tenenbaum,
  ``Neural-symbolic vqa: Disentangling reasoning from vision and language
  understanding,'' in \emph{NIPS}, 2018.

\bibitem{externVQA}
A.~Kumar, O.~Irsoy, P.~Ondruska, M.~Iyyer, J.~Bradbury, I.~Gulrajani, V.~Zhong,
  R.~Paulus, and R.~Socher, ``Ask me anything: Dynamic memory networks for
  natural language processing,'' in \emph{ICML}, 2016.

\bibitem{Narasimhan_2018_ECCV}
M.~Narasimhan and A.~G. Schwing, ``Straight to the facts: Learning knowledge
  base retrieval for factual visual question answering,'' in \emph{ECCV}, 2018.

\bibitem{zhu2017cvpr}
Y.~Zhu, J.~J. Lim, and L.~Fei-Fei, ``{Knowledge Acquisition for Visual Question
  Answering via Iterative Querying},'' in \emph{{CVPR}}, 2017.

\bibitem{Misra_2018_CVPR}
I.~Misra, R.~Girshick, R.~Fergus, M.~Hebert, A.~Gupta, and L.~van~der Maaten,
  ``Learning by asking questions,'' in \emph{CVPR}, 2018.

\bibitem{IQA}
D.~Gordon, A.~Kembhavi, M.~Rastegari, J.~Redmon, D.~Fox, and A.~Farhadi, ``Iqa:
  Visual question answering in interactive environments,'' in \emph{CVPR},
  2018.

\bibitem{Li_2018_CVPR}
Y.~Li, N.~Duan, B.~Zhou, X.~Chu, W.~Ouyang, X.~Wang, and M.~Zhou, ``Visual
  question generation as dual task of visual question answering,'' in
  \emph{CVPR}, 2018.

\bibitem{Teney_2018_ECCV}
D.~Teney and A.~van~den Hengel, ``Visual question answering as a meta learning
  task,'' in \emph{ECCV}, 2018.

\bibitem{narasimhan2018out}
M.~Narasimhan, S.~Lazebnik, and A.~G. Schwing, ``Out of the box: Reasoning with
  graph convolution nets for factual visual question answering,'' in
  \emph{NIPS}, 2018.

\bibitem{ECCV16baseline}
A.~Jabri, A.~Joulin, and L.~van~der Maaten, ``Revisiting visual question
  answering baselines,'' in \emph{ECCV}, 2016.

\bibitem{CountQA}
A.~Agrawal, D.~Batra, D.~Parikh, and A.~Kembhavi, ``Don't just assume; look and
  answer: Overcoming priors for visual question answering,'' in \emph{CVPR},
  2018.

\bibitem{DVQA}
K.~Kafle, B.~Price, S.~Cohen, and C.~Kanan, ``Dvqa: Understanding data
  visualizations via question answering,'' in \emph{CVPR}, 2018.

\bibitem{webchild}
N.~Tandon, G.~{de Melo}, F.~M. Suchanek, and G.~Weikum, ``{WebChild}:
  Harvesting and organizing commonsense knowledge from the web,'' in
  \emph{WSDM}, B.~Carterette, F.~Diaz, C.~Castillo, and D.~Metzler, Eds.\hskip
  1em plus 0.5em minus 0.4em\relax {ACM}, 2014.

\bibitem{conceptnet}
R.~Speer, J.~Chin, and C.~Havasi, ``Conceptnet 5.5: An open multilingual graph
  of general knowledge,'' in \emph{AAAI}, 2017.

\bibitem{dbpedia}
S.~Auer, C.~Bizer, G.~Kobilarov, J.~Lehmann, R.~Cyganiak, and Z.~Ives,
  ``Dbpedia: A nucleus for a web of open data,'' in \emph{ISWC+ASWC}, 2008.

\bibitem{mfh}
Z.~Yu, J.~Yu, C.~Xiang, J.~Fan, and D.~Tao, ``Beyond bilinear: Generalized
  multi-modal factorized high-order pooling for visual question answering,''
  \emph{IEEE TNNLS}, vol.~29, no.~12, pp. 5947--5959, 2018.

\bibitem{buattention}
P.~Anderson, X.~He, C.~Buehler, D.~Teney, M.~Johnson, S.~Gould, and L.~Zhang,
  ``Bottom-up and top-down attention for image captioning and visual question
  answering,'' in \emph{CVPR}, 2018.

\bibitem{tips}
D.~Teney, P.~Anderson, X.~He, and A.~van~den Hengel, ``Tips and tricks for
  visual question answering: Learnings from the 2017 challenge,'' in
  \emph{CVPR}, 2018.

\bibitem{rotate}
Z.~Sun, Z.-H. Deng, J.-Y. Nie, and J.~Tang, ``Rotate: Knowledge graph embedding
  by relational rotation in complex space,'' in \emph{ICLR}, 2018.

\bibitem{transe}
A.~Bordes, N.~Usunier, A.~Garcia-Duran, J.~Weston, and O.~Yakhnenko,
  ``Translating embeddings for modeling multi-relational data,'' in
  \emph{NIPS}, 2013.

\end{thebibliography}

\begin{IEEEbiography}[{\includegraphics[width=1in,height=1.25in,clip,keepaspectratio]{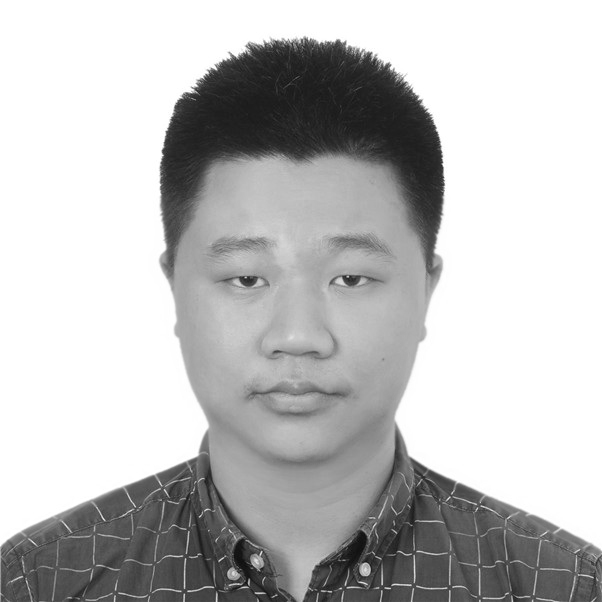}}]{Qingxing Cao}
	is currently a postdoctoral researcher in the School of Intelligent Systems Engineering at Sun Yat-sen University, working with Prof. Xiaodan Liang. He received his Ph.D. degree from Sun Yat-Sen University in 2019, advised by Prof. Liang Lin. His current research interests include computer vision and visual question answering.
\end{IEEEbiography}
\begin{IEEEbiography}[{\includegraphics[width=1in,height=1.25in,clip,keepaspectratio]{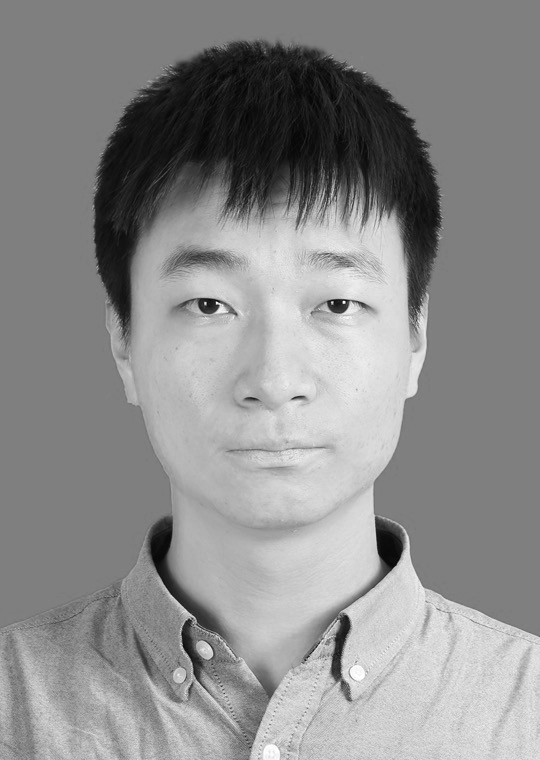}}]{Bailin Li}
	received his B.E. degree from Jilin University, Changchun, China, in 2016, and the M.S. degree at Sun Yat-Sen University, Guangzhou, China, advised by Professor Liang Lin. He currently leads the model optimization team at DMAI. His current research interests include visual reasoning and deep learning (e.g., network pruning, neural architecture search).
\end{IEEEbiography}
\begin{IEEEbiography}[{\includegraphics[width=1in,height=1.25in,clip,keepaspectratio]{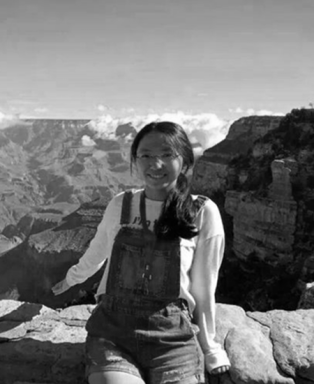}}]{Xiaodan Liang}
	is currently an Associate Professor at Sun Yat-sen University. She was a postdoc researcher in the machine learning department at Carnegie Mellon University, working with Prof. Eric Xing, from 2016 to 2018. She received her PhD degree from Sun Yat-sen University in 2016, advised by Liang Lin. She has published several cutting-edge projects on human-related analysis, including human parsing, pedestrian detection and instance segmentation, 2D/3D human pose estimation and activity recognition.
\end{IEEEbiography}
\begin{IEEEbiography}[{\includegraphics[width=1in,height=1.25in,clip,keepaspectratio]{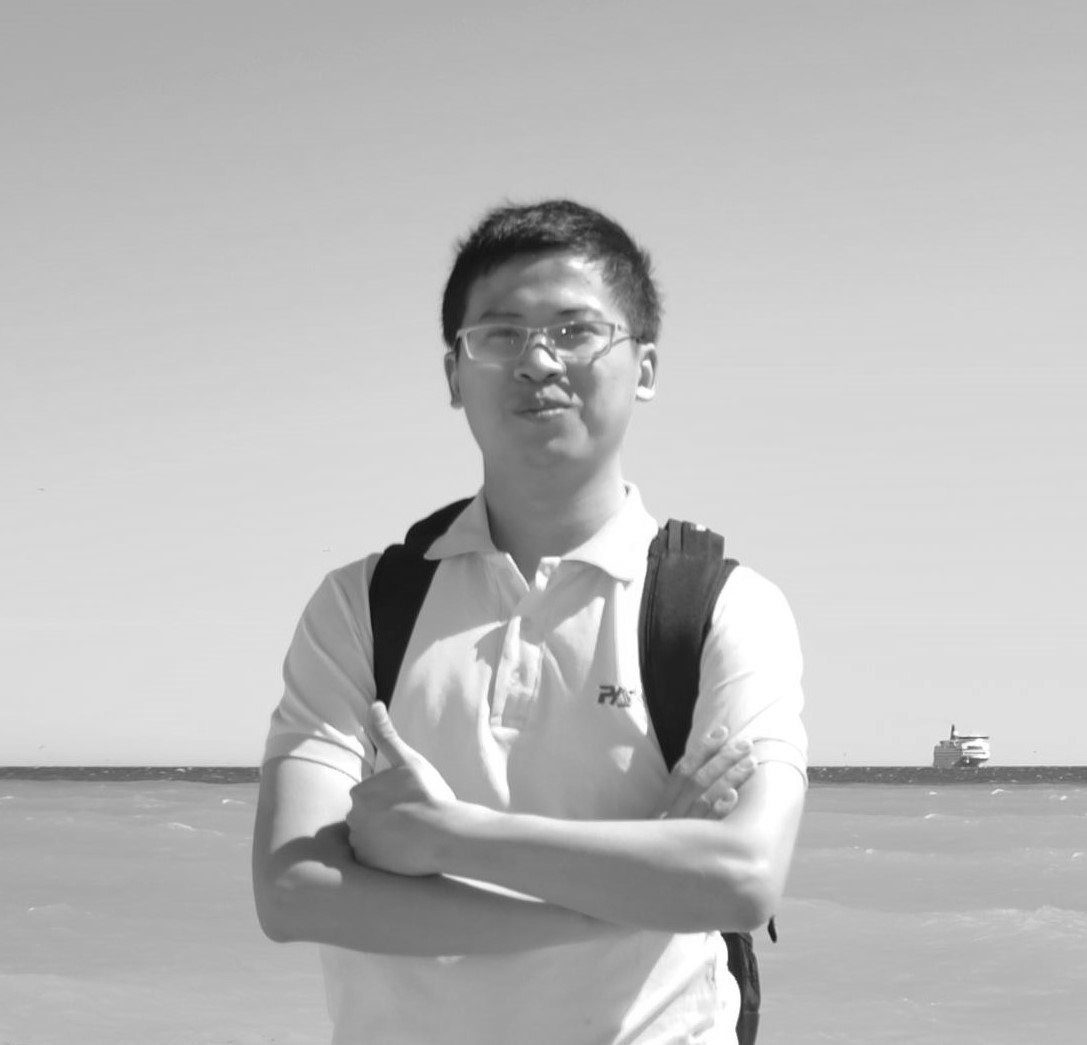}}]{Keze Wang}
	received his B.S. degree in software engineering from Sun Yat-Sen University, Guangzhou, China, in 2012. He is currently pursuing his dual Ph.D. degree at Sun Yat-Sen University and Hong Kong Polytechnic University, advised by Prof. Liang Lin and Lei Zhang. His current research interests include computer vision and machine learning. More information can be found on his personal website http://kezewang.com.
\end{IEEEbiography}
\begin{IEEEbiography}[{\includegraphics[width=1in,height=1.25in,clip,keepaspectratio]{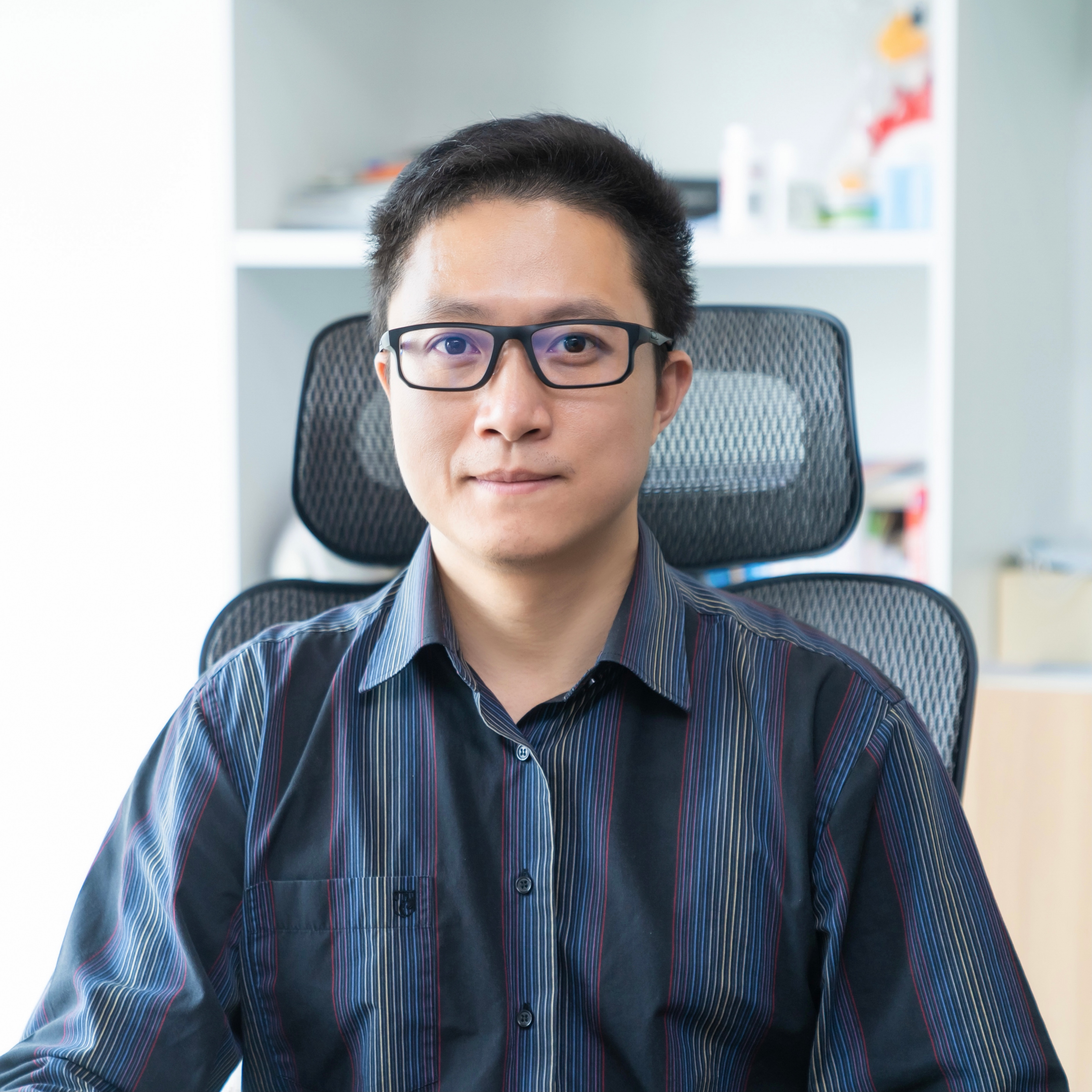}}]{Liang Lin}
	is a full professor of Computer Science in Sun Yat-sen University and CEO of DarkerMatter AI. He worked as the Executive Director of the SenseTime Group from 2016 to 2018, leading the R\&D teams in developing cutting-edge, deliverable solutions in computer vision, data analysis and mining, and intelligent robotic systems.  He has authored or co-authored more than 200 papers in leading academic journals and conferences. He is an associate editor of IEEE Trans.  Human-Machine Systems and IET Computer Vision, and he served as the area/session chair for numerous conferences such as CVPR, ICME, ICCV. He was the recipient of Annual Best Paper Award by Pattern Recognition (Elsevier) in 2018, Dimond Award for best paper in IEEE ICME in 2017, ACM NPAR Best Paper Runners-Up Award in 2010, Google Faculty Award in 2012, award for the best student paper in IEEE ICME in 2014, and Hong Kong Scholars Award in 2014. He is a Fellow of IET.
\end{IEEEbiography}
\end{document}